\documentclass[10pt,twocolumn,letterpaper]{article}
\usepackage{algorithm}
\usepackage{algpseudocode}
\usepackage{graphicx}
\usepackage{amsmath}
\usepackage{amssymb}
\usepackage{booktabs} 
\usepackage{tcolorbox}
\usepackage[dvipsnames]{xcolor}
\usepackage{pifont}
\usepackage{tabularx}
\usepackage[dvipsnames]{xcolor}
\newcommand{\cmark}{{\color{ForestGreen}\ding{51}}}
\newcommand{\xmark}{{\color{red}\ding{55}}}
\usepackage[pagenumbers]{cvpr}              
\definecolor{cvprblue}{rgb}{0.21,0.49,0.74}
\usepackage[pagebackref,breaklinks,colorlinks,allcolors=cvprblue]{hyperref}

\def\paperID{16158} 
\def\confName{CVPR}
\def\confYear{2026}

\title{Edit-As-Act: Goal-Regressive Planning for Open-Vocabulary 3D Indoor Scene Editing}

\author{
Seongrae Noh \quad
SeungWon Seo \quad
Gyeong-Moon Park$^{\dagger}$ \quad
HyeongYeop Kang$^{\dagger}$ \\[1.2em]
Korea University\\[0.3em]
\texttt{\{rhosunr99, ssw03270, gm-park, siamiz\_hkang\}@korea.ac.kr
}\\[0.5em]
\url{https://seongraenoh.github.io/edit-as-act/}
}

\begin{document}

\maketitle
\begingroup
\renewcommand{\thefootnote}{\fnsymbol{footnote}}
\footnotetext[2]{Corresponding authors.}
\endgroup

\begin{abstract}
Editing a 3D indoor scene from natural language is conceptually straightforward but technically challenging. Existing open-vocabulary systems often regenerate large portions of a scene or rely on image-space edits that disrupt spatial structure, resulting in unintended global changes or physically inconsistent layouts. These limitations stem from treating editing primarily as a generative task.
We take a different view. A user instruction defines a desired world state, and editing should be the minimal sequence of actions that makes this state true while preserving everything else. This perspective motivates Edit-As-Act, a framework that performs open-vocabulary scene editing as goal-regressive planning in 3D space.
Given a source scene and free-form instruction, Edit-As-Act predicts symbolic goal predicates and plans in EditLang, a PDDL-inspired action language that we design with explicit preconditions and effects encoding support, contact, collision, and other geometric relations. A language-driven planner proposes actions, and a validator enforces goal-directedness, monotonicity, and physical feasibility, producing interpretable and physically coherent transformations.
By separating reasoning from low-level generation, Edit-As-Act achieves instruction fidelity, semantic consistency, and physical plausibility—three criteria that existing paradigms cannot satisfy together. On E2A-Bench, our benchmark of 63 editing tasks across 9 indoor environments, Edit-As-Act significantly outperforms prior approaches across all edit types and scene categories.
\end{abstract}
    
\section{Introduction}
\label{sec:intro}
Training and evaluating embodied agents increasingly depends on environments that can be modified with precision. While recent models can synthesize open-vocabulary indoor scenes with remarkable diversity, their usefulness is limited without an equally reliable way to edit those scenes in response to high-level goals. Tasks such as rearranging furniture, preparing simulation curricula, or adapting a space for downstream interaction require more than scene generation. They require transformations that are intentional, localized to the relevant region, and physically valid.

Achieving this level of control requires meeting three criteria at once. The edit must follow the instructions. It must preserve the rest of the scene. It must remain physically plausible. As shown in~\cref{tab:single_col_comparison_better}, representative systems from layout generation, constraint optimization, and image-based editing each satisfy only part of this triad. They often drift semantically, modify regions that should remain unchanged, or violate geometric constraints. This pattern indicates that these limitations arise from the underlying paradigms rather than from specific implementations.

Motivated by both embodied agents, which learn through sequential decision-making, and classical automated planning~\cite{fikes1971strips,geffner2013concise,haslum2019introduction, verma2025teaching}, which frames tasks as goal satisfaction through structured actions, we view scene editing as an inherently goal-driven process. A natural-language instruction implicitly defines a desired world state. The correct edit is the minimal sequence of actions that achieves this state while remaining consistent with the geometry of the original scene. This joint perspective from embodied intelligence and symbolic planning naturally leads to a backward formulation, where reasoning starts from the desired outcome and works in reverse to identify the necessary actions and their preconditions.

We propose \textit{Edit-As-Act}, a framework that performs open-vocabulary 3D scene editing through goal-regressive reasoning. Natural-language instructions are mapped to symbolic goal predicates expressed in EditLang, a PDDL-inspired editing language that we design specifically for this task.
It defines atomic edit actions through explicit preconditions and effects. Two large language model-driven modules then carry out backward planning: a planner proposes actions that satisfy current goals, while a validator enforces goal-directedness, monotonicity, and geometric feasibility. Unmet preconditions recursively generate subgoals, and the process continues until all requirements hold in the source scene. We formulate 3D scene editing as goal satisfaction over state transitions rather than global re-generation, enabling executable and verifiable editing through source-aware regression.

By grounding edits in symbolic structure and verifiable geometry, Edit-As-Act produces interpretable, physically consistent, and semantically faithful transformations beyond what generative or optimization-based methods can ensure. To assess these capabilities, we introduce \textit{E2A-Bench}, a set of 63 open-vocabulary editing tasks across 9 indoor environments that measure instruction alignment, semantic stability, and physical realism.

Our contributions can be summarized as follows:
\begin{itemize}
\item \textbf{Reasoning-driven editing framework}: Introducing \textit{Edit-As-Act}, the first framework to cast open-vocabulary 3D scene editing as a sequential, goal-regressive reasoning problem.
\item \textbf{Verifiable symbolic action language}: Developing \textit{EditLang}, a symbolic editing language with explicit preconditions and effects that guarantee logical coherence and spatial validity.
\item \textbf{Open-vocabulary benchmark}: Creating \textit{E2A-Bench}, a suite of 63 editing tasks across 9 indoor environments for standardized evaluation of semantic alignment, edit reliability, and physical realism.
\end{itemize}

\begin{table}[t]
  \centering
  \small
  \caption{Comparison of three essential requirements for different scene editing paradigms. LayoutGPT represents direct 3D layout editing. AnyHome represents constraint-based optimization. ArtiScene represents image-driven editing followed by 3D lifting.}
  \label{tab:single_col_comparison_better}
  \begin{tabular*}{\linewidth}{l@{\extracolsep{\fill}}ccc}
    \toprule
    & Instruction & Semantic & Physical \\
    & Fidelity & Consistency & Plausibility \\
    \midrule
    LayoutGPT~\cite{feng2023layoutgpt}    & \cmark & \xmark & \xmark \\
    AnyHome~\cite{fu2024anyhome}     & \xmark & \xmark & \cmark \\
    ArtiScene~\cite{gu2025artiscene}   & \xmark & \cmark & \xmark \\
    \midrule
    \textbf{Edit-As-Act} (Ours) & \cmark & \cmark & \cmark \\
    \bottomrule
  \end{tabular*}
\end{table}

\section{Related Work}
\label{sec:related_work}

\subsection{Data Driven Scene Layout Editing}
Many approaches perform scene editing using data-driven generators trained on layout or scene graph datasets such as 3D-FRONT~\cite{fu20213d}. Methods including DiffuScene~\cite{tang2024diffuscene}, InstructScene~\cite{lin2024instructscene}, EditRoom~\cite{zheng2024editroom}, and EchoScene~\cite{zhai2024echoscene} diffuse over discrete scene structures to produce text-conditioned layouts. These models generate visually plausible edits when instructions remain within the training distribution, but their performance degrades for novel rooms or compositional queries. Because edits occur through a single forward generative step, local modifications can induce unintended global changes. Physical validity is also only implicit; for instance, EditRoom~\cite{zheng2024editroom} reports that erroneous language model commands may produce colliding or unsupported objects.

\subsection{Constraint-Based Scene Layout Synthesis}
Other methods convert language into spatial relations or optimization objectives. Earlier work mapped descriptions to fixed scene graphs for retrieval~\cite{ma2018language}. Recent systems such as Holodeck~\cite{yang2024holodeck}, I-Design~\cite{ccelen2024design}, LLPlace~\cite{yang2024llplace}, AnyHome~\cite{fu2024anyhome}, and LayoutVLM~\cite{sun2025layoutvlm} use language models to derive placement rules or constraints and then optimize layouts to satisfy them. These approaches are physically grounded but often re-optimize whole rooms, causing non-target objects to shift and reducing scene-level consistency~\cite{yang2024holodeck, ccelen2024design}. When constraints conflict, solvers may settle for partially satisfied solutions, lowering instruction fidelity. LLM-based planners such as LayoutGPT~\cite{feng2023layoutgpt} also struggle with 3D spatial reasoning, leading to misaligned placements.

\subsection{2D-to-3D Image-Based Editing}
A third direction edits scenes by generating an instruction-consistent image and lifting it to 3D. ArtiScene~\cite{gu2025artiscene}, Text2Room~\cite{hollein2023text2room}, and SceneScape~\cite{fridman2023scenescape} reconstruct geometry from edited images, while ControlRoom3D~\cite{schult2024controlroom3d} and Ctrl-Room~\cite{fang2025ctrl} add geometric guidance. These approaches inherit strong 2D priors but lack explicit 3D reasoning, often producing structural artifacts or physically implausible layouts~\cite{gu2025artiscene, hollein2023text2room}. Edits made in image space are also difficult to localize cleanly and can introduce unintended changes in unedited regions.

Generative diffusion, constraint-based optimization, and image-driven lifting each contribute valuable insight to language-conditioned scene editing, but all face challenges: reliance on training data manifolds, difficulty maintaining edit locality, and limited explicit reasoning about 3D structure and physics. Our work addresses these limitations by formulating scene editing as goal-regressive planning in a symbolic action space. 
\section{Problem Definition}
\label{sec:problem_definition}

\begin{figure*}[t]
\centering
\includegraphics[width=1\linewidth]{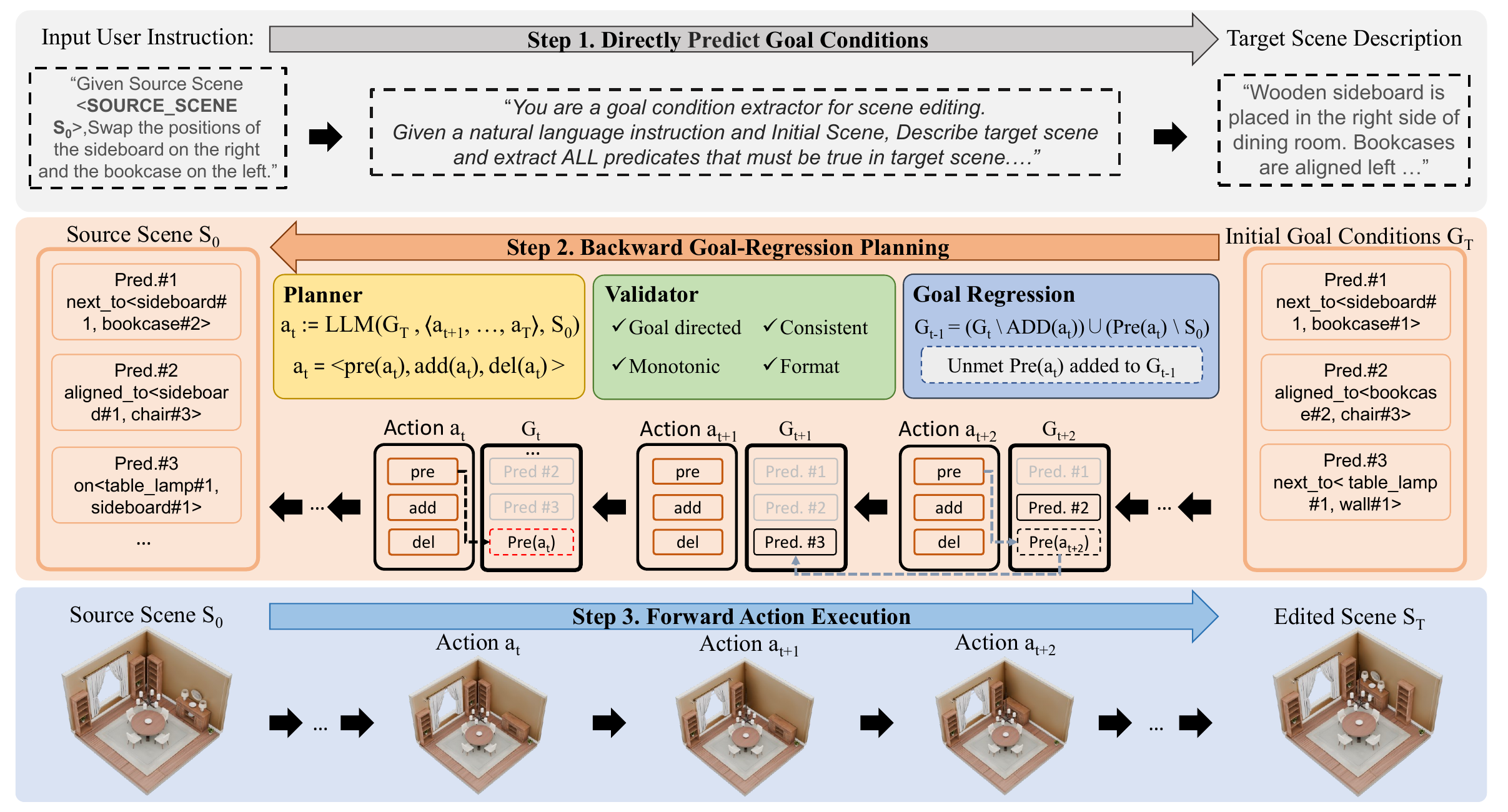}
\caption{Overview of \textbf{Edit-As-Act}. 
\textbf{Step 1}: an LLM converts a source scene $S_0$ and instruction into symbolic goal predicates $G_T$ in EditLang. 
\textbf{Step 2}: a planner–validator loop iteratively selects EditLang actions $a_t$ that satisfy goals $G_t$ and regresses remaining goals until all are grounded in $S_0$. 
\textbf{Step 3}: the resulting action sequence is applied to $S_0$ to obtain the edited scene $S_T$.}
\label{fig:overview}
\end{figure*}

We reformulate open-vocabulary indoor scene editing as a goal-regressive planning problem inspired by classical STRIPS reasoning~\cite{fikes1971strips,geffner2013concise,haslum2019introduction}. 
This formulation turns free-form language instructions into verifiable symbolic goals, defining an editing problem through the symbolic vocabulary, available actions, initial scene, and desired post-edit conditions.

We use a PDDL-style~\cite{aeronautiques1998pddl} vocabulary of predicates and actions. Let $\mathcal{F}$ be the finite set of ground predicates describing facts in the 3D world (e.g., $\texttt{on}(\texttt{chair},\texttt{floor})$ or $\texttt{facing}(\texttt{chair},\texttt{table})$). A symbolic state $s$ is any subset of $\mathcal{F}$, with the source scene encoded as $s_0$. Goal conditions $G \subseteq \mathcal{F}$ are derived from the instruction $I$. 
The available edits form a finite action set $\mathcal{A}$, where each action $a$ is a triplet $\langle \mathrm{pre}(a), \mathrm{add}(a), \mathrm{del}(a) \rangle$. Preconditions specify when $a$ is executable, $\mathrm{add}(a)$ describes predicates it establishes, and $\mathrm{del}(a)$ removes incompatible ones. Mutually exclusive relations are automatically removed (for example, removing $\texttt{on}(x,y)$ before adding $\texttt{on}(x,z)$). Executing $a$ updates the state as
\begin{equation}
 s' = (s \setminus \mathrm{del}(a)) \cup \mathrm{add}(a).
\end{equation}
The goal is to find a sequence $\pi = \langle a_1,\dots,a_T\rangle$ that transforms $s_0$ into a target state satisfying all goals in $G$.

Classical STRIPS regression~\cite{geffner2013concise, haslum2019introduction} works backward from the goal $G$ to derive the minimal subgoals. However, directly applying it to 3D scenes causes redundant reasoning because many goals are already satisfied in $s_0$
We therefore introduce a source-aware regression operator that propagates only unsatisfied conditions:
\begin{equation}
\label{eq:source aware regression operator}
\mathrm{Regress}^*(G,a; s_0)=\bigl(G\setminus \mathrm{add}(a)\bigr)\ \cup\ \bigl(\mathrm{pre}(a)\setminus s_0\bigr).
\end{equation}
This preserves the logical rigor of STRIPS while avoiding unnecessary reconstruction of scene aspects already satisfied in the source, forming the formal basis for our method.

\section{Method}

\subsection{Framework Overview}

Edit-As-Act performs open-vocabulary three dimensional scene editing as a process of \emph{goal-regressive reasoning} rather than direct generation. Given a natural language instruction and a source scene, the system constructs a verifiable sequence of edit actions that transforms the source into a target state satisfying all goal conditions. Two large language model based modules, a \emph{planner} and a \emph{validator}, operate within our EditLang domain to propose actions, regress goals, and check logical and physical validity. Before planning, we evaluate all EditLang predicates on the source scene $S_0$ under a closed world assumption to obtain the initial symbolic state $s_0$. Conceptually, actions update symbolic states via the STRIPS transition $s' = (s \setminus \mathrm{del}(a)) \cup \mathrm{add}(a)$ defined in \cref{sec:problem_definition}, while in our implementation we recompute predicates from the updated geometry after each accepted edit to keep symbols and the three dimensional scene aligned. This backward, source aware process yields minimal edits, preserves semantic context and physical plausibility, and avoids layout hallucination. Our framework is summarized in ~\cref{fig:overview}


\subsection{EditLang}
EditLang provides the symbolic foundation of Edit-As-Act. It defines a PDDL-style domain tailored for open-vocabulary scene editing, consisting of \emph{predicates} and \emph{actions} that bridge language instructions and geometric reasoning. Predicates capture geometric, topological, and physical relations such as $\texttt{supported}(x,y)$, $\texttt{contact}(x,y)$, $\texttt{clear}(x)$, $\texttt{stable}(x)$, $\texttt{colliding}(x,y)$, and $\texttt{reachable}(x)$, all evaluated directly from $s_0$. 

Actions are atomic and deterministic, defined as $\langle \mathrm{pre},\mathrm{add},\mathrm{del}\rangle$ introduced in \cref{sec:problem_definition}. EditLang uses the same symbolic vocabulary across tasks, but unlike traditional benchmark planning domains with a fixed, hand designed object set, it is instantiated \emph{per scene} by dynamically binding typed variables to concrete objects in the source scene. This instantiation allows the domain to reason over unseen object categories and layouts while keeping the set of ground predicates finite. All objects and their attributes are registered and accessible at plan time. 
Beyond geometric rearrangement, EditLang also supports two non geometric edit primitives, \texttt{Add} and \texttt{Stylize}. \texttt{Add} inserts assets from a catalog associated with the source scene into collision free, supported locations that respect room specific constraints. \texttt{Stylize} modifies appearance level attributes such as material or color by updating an object's description, while leaving its geometric configuration unchanged. Together with the rearrangement actions, these primitives cover position, existence, and appearance edits in a single symbolic language. See Supplementary Sec.~4 for a formalization of EditLang.


\subsection{Planner Module}
The planner $P$ translates an instruction $I$ and source scene $S_0$ into an initial goal predicate set $G_T$ written in EditLang. At each step $t$, $P$ receives $G_t$, $s_0$, and the partial plan $\langle a_1,\dots,a_{t-1}\rangle$ and proposes a single action
\[
a_t=\langle \mathrm{pre}(a_t),\, \mathrm{add}(a_t),\, \mathrm{del}(a_t)\rangle
\]
that satisfies at least one goal in $G_t$.
We prompt $P$ to produce minimal but sufficient preconditions $\mathrm{pre}(a_t)$ that guarantee physical executability across scenes. This is enforced through geometric checks for collision, support, and stability with fixed numeric tolerances in scene units. The planner submits $a_t$ to the validator and only revises the proposal in response to validator feedback; in practice, we cap the number of revisions per step at three. Please see Supplementary Sec.~5 for implementation details and pseudo code of the Planner module.

\subsection{Validator Module}
The validator $V$ evaluates each proposed action $a_t$ on four criteria that directly operationalize our desiderata in \cref{sec:intro}.

\emph{Goal directedness.} The added predicates make progress on the current goals: $\mathrm{add}(a_t)$ must satisfy at least one element of $G_t$. This prevents tangential edits that do not contribute to the instruction.

\emph{Monotonicity.} The action must not undo progress on goals that have already been achieved. Let $G_{\le t}^{\mathrm{sat}}$ denote the subset of goals satisfied before step $t$; $V$ enforces $\mathrm{del}(a_t)\cap G_{\le t}^{\mathrm{sat}}=\varnothing$. Together with the finite EditLang state space, this monotonicity constraint rules out cycles and guarantees that the regression loop terminates in finitely many steps.

\emph{Contextual consistency.} When the first two criteria hold, $V$ checks that the resulting configuration remains plausible with respect to the source scene and room specific constraints (for example, seating around a table or clearance in front of doors), capturing semantic coherence beyond a pure checklist of predicates.

\emph{Formal validity.} Finally, $V$ verifies that the action conforms to the EditLang schema: predicate and argument types are well formed, mutually exclusive relations are updated consistently, and action names and parameters match the domain specification.

We maintain a set of domain invariants $\mathcal{I}$ that includes collision freedom, a single stable support per object, and wall or floor attachment rules; $V$ rejects any action that violates $\mathcal{I}$, even if it is otherwise goal directed. On failure, $V$ returns a refusal with a brief natural language explanation, which is fed back to the planner. On success, $V$ accepts $a_t$ and passes it to the regression step. See Supplementary Sec.~6 for implementation details of the Validator module and additional failure cases.

\subsection{Source Aware Goal Regression}
After acceptance, goals are updated by source-aware regression with the source scene $S_0$,
\[
G_{t-1}=\bigl(G_t \setminus \mathrm{add}(a_t)\bigr)\ \cup\ \bigl(\mathrm{pre}(a_t) \setminus S_0\bigr).
\]

We assume disjoint effects $\mathrm{add}(a)\cap \mathrm{del}(a)=\varnothing$ and consistent variable bindings and remove mutually exclusive predicates via $\mathrm{del}$ before applying $\mathrm{add}$. This preserves the logic of classical STRIPS regression while avoiding reconstruction of parts of the scene that are already satisfied in the source state. The planner then proceeds with the updated goal set $G_{t-1}$.

\subsection{Execution and Termination}
The planning process continues until $G_t=\emptyset$, indicating that all goal conditions have been regressed back to the source scene. The backward plan $\langle a_T,\dots,a_1\rangle$ is then reversed and executed by a deterministic Python DSL runtime that invokes the corresponding implementation for each EditLang action. After each invocation, predicates are recomputed from geometry before proceeding to the next step, ensuring that the symbolic state and 3D scene remain consistent. Because EditLang induces a finite set of ground predicates over the objects of the source scene and the validator rules out cyclic or non-monotone updates, the overall planning loop always terminates in a finite number of steps. Examples of planning scenarios, the full EditLang specification, and the DSL implementations are provided in the supplementary material.
\section{Experiments}
\label{sec:experiment}

\begin{table*}[!ht]
\centering
\caption{Quantitative comparison across nine scene categories. Edit-As-Act achieves the strongest and most consistent performance in instruction fidelity (IF), semantic consistency (SC), and physical plausibility (PP), demonstrating robust generalization across diverse spatial configurations.}
\centering
\small
\label{tab:benchmark_scene_category}

\resizebox{\textwidth}{!}{%
\begin{tabular}{lccc ccc ccc ccc ccc}
\toprule
& \multicolumn{3}{c}{Bathroom} & \multicolumn{3}{c}{Bedroom} & \multicolumn{3}{c}{Computer Room} & \multicolumn{3}{c}{Dining Room} & \multicolumn{3}{c}{Game Room} \\
\cmidrule(lr){2-4} \cmidrule(lr){5-7} \cmidrule(lr){8-10} \cmidrule(lr){11-13} \cmidrule(lr){14-16}
Methods & IF & SC & PP & IF & SC & PP & IF & SC & PP & IF & SC & PP & IF & SC & PP \\
\midrule
LayoutGPT-E & 50.1 & 62.5 & 78.3 & 35.7 & 30.2 & 53.8 & 53.8 & 32.1 & 84.7 & 50.5 & 31.4 & 66.2 & 52.1 & 48.9 & 83.5 \\
AnyHome  & \textbf{69.7} & 56.7 & 81.7 & \textbf{64.0} & 62.6 & 82.7 & 59.0 & 56.3 & 90.7 & 57.1 & 66.9 & 76.6 & 58.3 & 67.1 & \textbf{91.4} \\
ArtiScene-E  & 61.7 & 73.4 & 83.7 & 43.0 & 39.0 & 90.4 & 41.9 & 41.0 & 90.6 & 37.3 & 37.1 & 90.4 & \textbf{61.3} & 54.4 & 89.4 \\
\textbf{Edit-As-Act} (ours) & 58.7 & \textbf{88.9} & \textbf{89.1} & 45.7 & \textbf{73.1} & \textbf{91.9} & \textbf{73.6} & \textbf{88.0} & \textbf{94.1} & \textbf{89.7} & \textbf{95.3} & \textbf{92.7} & 58.7 & \textbf{79.9} & 91.1 \\
\bottomrule
\end{tabular}%
}

\vspace{1.5ex} 

\resizebox{\textwidth}{!}{%
\begin{tabular}{lccc ccc ccc ccc ccc}
\toprule
& \multicolumn{3}{c}{Kids Room} & \multicolumn{3}{c}{Kitchen} & \multicolumn{3}{c}{Living Room} & \multicolumn{3}{c}{Office} & \multicolumn{3}{c}{Average} \\
\cmidrule(lr){2-4} \cmidrule(lr){5-7} \cmidrule(lr){8-10} \cmidrule(lr){11-13} \cmidrule(lr){14-16}
Methods &IF &SC &PP & IF & SC & PP & IF & SC & PP & IF & SC & PP & IF & SC & PP \\
\midrule
LayoutGPT-E & 28.3 & 30.1 & 88.6 & 59.4 & 58.2 & 88.3 & 38.6 & 40.7 & 83.9 & 32.1 & 45.4 & 77.1 & 42.3 & 48.8 & 78.6 \\
AnyHome  & 72.0 & 82.0 & 92.7 & 49.3 & 51.0 & 82.9 & 44.7 & 58.1 & 81.6 & 44.6 & 43.9 & 80.4 & 57.6 & 60.5 & 84.5 \\
ArtiScene-E  & 35.0 & 37.1 & 92.4 & \textbf{65.6} & 67.6 & \textbf{94.1} & 48.9 & 68.1 & 89.1 & 40.0 & 42.9 & \textbf{92.9} & 48.3 & 51.2 & 90.3 \\
\textbf{Edit-As-Act} (ours) & \textbf{91.1} & \textbf{89.0} & \textbf{96.3} & 55.0 & \textbf{92.3} & 93.7 & \textbf{72.9} & \textbf{90.1} & \textbf{93.6} & \textbf{76.4} & \textbf{82.9} & 81.9 & \textbf{69.1} & \textbf{86.6} & \textbf{91.7} \\
\bottomrule
\end{tabular}%
}
\end{table*}

\subsection{Benchmark Setup}
We evaluate our Edit-As-Act on an open-vocabulary indoor editing benchmark designed to test reasoning rather than pattern matching. The benchmark contains 63 editing tasks across 9 diverse indoor environments (bathrooms, bedrooms, computer rooms, dining rooms, game rooms, kids' rooms, kitchens, living rooms, and offices). Each task provides a source 3D layout with geometry and object metadata (pose, category labels, front face orientation), along with a free-form language instruction. 
Instructions range from simple rearrangements (e.g., ``move the table near the sofa") to multi-step compositional edits (e.g., ``rotate the chair to face the window and place a lamp beside it"). 

Unlike prior evaluations that focus on reproducing target layouts, our benchmark is constructed to stress goal-conditioned reasoning, symbolic interpretability, and spatial generalization across unseen scene configurations. It tests whether a model can understand an instruction, identify the minimal required changes, and update the scene without disturbing unrelated context. See the supplementary material for details on the dataset generation pipeline.


\subsection{Baseline Methods and Metrics}
We compare Edit-As-Act against three state-of-the-art baselines, each representing a different editing paradigm.
\textbf{LayoutGPT-E}~\cite{feng2023layoutgpt} performs forward reasoning in layout space. We provide the source layout directly in the prompt. Then, the model outputs an edited layout. For add and stylize operations, it is further prompted to describe the inserted or modified objects so that the edited layout can be reconstructed consistently.
\textbf{AnyHome}~\cite{fu2024anyhome} is a constraint-based optimization framework that converts language instructions into symbolic spatial constraints and solves for a layout that satisfies them. Because AnyHome is originally designed for multi-room floorplan synthesis, we disable all modules that operate across rooms.
\textbf{ArtiScene-E}~\cite{gu2025artiscene} represents an image first editing pipeline. The source scene is rendered with fixed camera and lighting. A pretrained text-to-image model generates an edited image conditioned on the instruction. ArtiScene then lifts this edited image into a structured 3D scene. 

We evaluate all methods with three complementary metrics that assess different aspects of editing quality:
\begin{enumerate}
    \item \textbf{Instruction Fidelity (IF)}: An LVLM-based score that measures how faithfully the edited scene satisfies the explicit instruction.
    \item \textbf{Semantic Consistency (SC)}: An LVLM-based score that evaluates whether the nontargeted regions of the scene remain unchanged.
    \item \textbf{Physical Plausibility (PP)}: An LVLM-based score that rates the overall physical plausibility, including collisions, support relations (e.g., objects resting on appropriate supports such as floors, tables, or shelves), and stability(e.g., furniture not floating or tipping over in implausible ways).
\end{enumerate}
Details of the prompts are provided in the supplementary material.

\subsection{Implementation Details}
We ensure a fair comparison by standardizing the model backbone, asset pipeline, and text-to-3D generation across all methods. LayoutGPT-E, AnyHome, and our planner–validator modules all rely on the OpenAI GPT-5 API~\cite{GPT-5} with identical decoding parameters. Whenever a method outputs a textual description for object addition or stylization, we generate the corresponding image with Gemini 2.5 Flash Image and convert it into consistent 3D geometry using Hyper3D Gen-2 V1.8~\cite{RodinGen-2}, identical to that used to construct the source scenes. For ArtiScene-E, the edited view is likewise produced using the same text-to-image interface before being lifted. 

This unified pipeline ensures the performance difference arises not from external generative modules, but from each method's reasoning strategy, constraint handling, and editing mechanism. 
\subsection{Quantitative Evaluation}
Across all 63 editing tasks, Edit-As-Act achieves the strongest overall performance in IF, SC, and PP. Results are shown in \cref{tab:benchmark_scene_category}. 


\paragraph{Instruction Fidelity.} 
\cref{tab:benchmark_scene_category} reveals that editing instructions challenge all baselines. 
LayoutGPT-E struggles due to one-shot generation and often fails to react to multi-predicate instructions.
AnyHome performs moderately but sometimes satisfies constraints in its abstract graph while failing to align all object placements in 3D.
ArtiScene-E performs well on PP due to strong image priors, but its weaker instruction fidelity on IF and SC stems from the text-to-image stage, which often produces edits that are conservative and insufficiently responsive to the instruction, resulting in outputs that remain overly close to the input image. 

\paragraph{Semantic Consistency.}
As shown in~\cref{tab:benchmark_scene_category}, SC exceeds 85 to 95 across kitchens, living rooms, and bedrooms, while LayoutGPT-E and AnyHome degrade sharply in scenes with many small objects. This pattern reflects a major limitation of forward generative and constraint optimization approaches: small changes often propagate globally, leading to context drift. Edit-As-Act avoids this failure mode through localized symbolic reasoning, modifying only what the goal demands.

\begin{figure*}[t]
\centering
\includegraphics[width=1\linewidth]{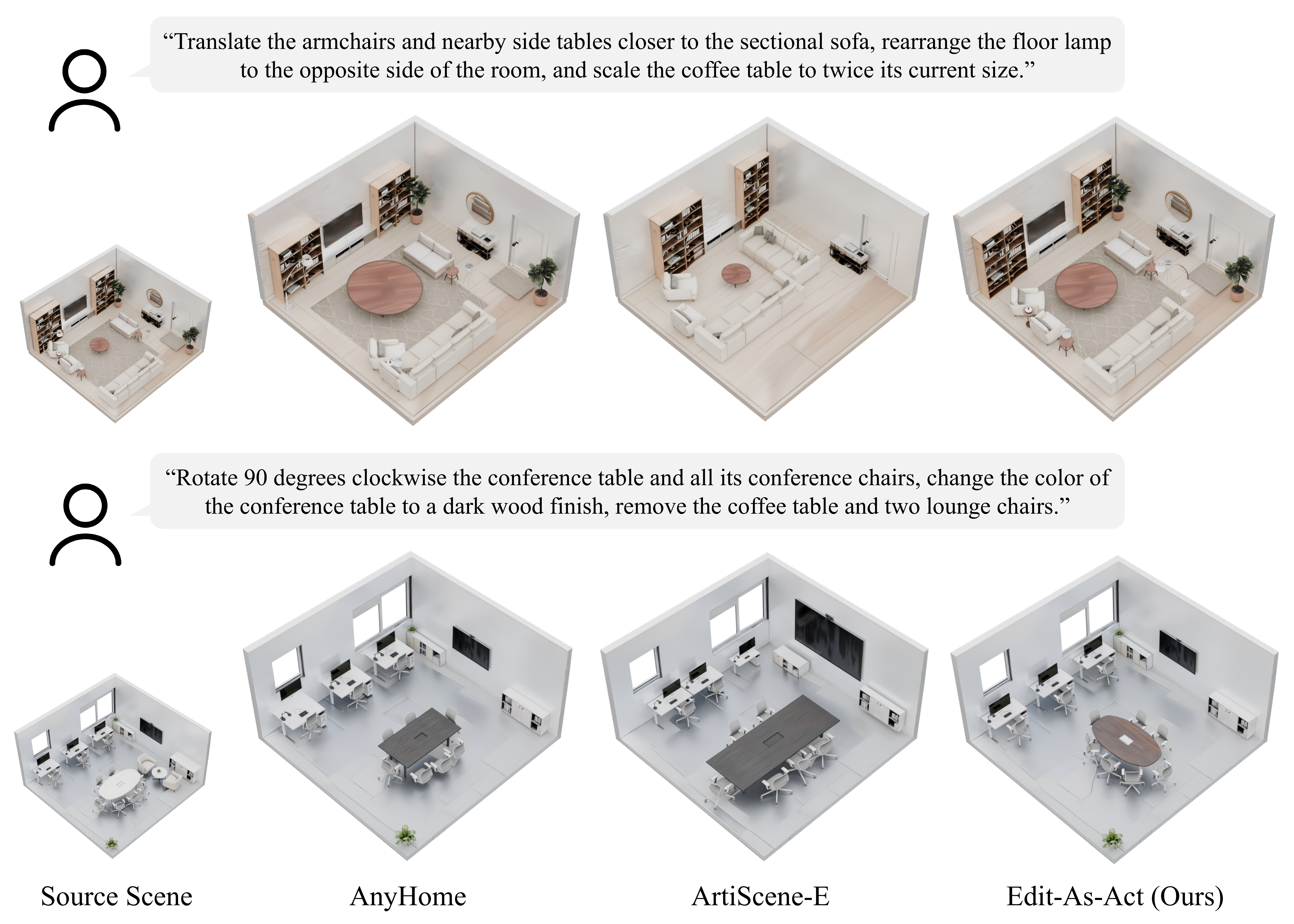}
\caption{Representative qualitative results. Baseline methods often introduce unintended global changes, fail to satisfy multi-step instructions, or generate incomplete edits. Edit-As-Act produces precise, instruction-aligned modifications that remain physically valid and preserve the overall scene identity.}
\label{fig:qualitative}
\end{figure*}

\paragraph{Physical Plausibility.} 
Edit-As-Act consistently delivers physically plausible configurations even for large geometric edits. 
In \cref{tab:benchmark_scene_category}, our PP scores remain near or above 90 across most categories, matching or outperforming baselines. Since every step in our plan is validated by explicit geometric checks, collisions and unstable placements are rarely introduced.

ArtiScene-E reports competitive PP because the generated 2D rendering often avoids visible collisions from the chosen viewpoint. However, these scores do not reflect full scene stability. When the edited render is lifted into 3D, many placements lack valid support or produce hidden interpenetrations that are not penalized by a single-view LVLM evaluator. As a result, ArtiScene's PP deteriorates in cluttered or occluded regions, where 2D edits fail to encode full geometric constraints.

Overall, the quantitative results reveal a clear structural pattern: forward generation struggles with compositional reasoning, constraint solving loses context during layout optimization, and image-to-3D lifting inherits ambiguity from the 2D model. By contrast, Edit-As-Act's goal-regressive reasoning provides a principled path that integrates instruction alignment, locality, and physical validity in a way that existing paradigms cannot. We provide a fine-grained analysis by edit operation type in supplementary material.


\begin{table}[t]
\centering
\caption{Comparison with contemporary reasoning baselines.}
\label{tab:additional_comparison}
\scriptsize
\setlength{\tabcolsep}{3.5pt}
\begin{tabular*}{\columnwidth}{@{\extracolsep{\fill}}lcccccc}
\toprule
Method & IF $\uparrow$ & SC $\uparrow$ & PP $\uparrow$ & Latency (s) & Avg. Calls & Avg. Tok \\
\midrule
GPT-5                & 49.6 & 52.3 & 73.3 & 18.9  & 1.0 & 3357  \\
Gemini-3-Pro-preview & 43.1 & 48.7 & 71.7 & 47.7  & 1.0 & 6959  \\
Claude-4.5-opus      & 50.2 & 43.5 & 68.5 & 11.3  & 1.0 & 3514  \\
SceneWeaver          & 68.7 & 78.3 & 82.1 & 102.5 & 7.3 & 18335 \\
\textbf{Edit-As-Act (ours)} & \textbf{69.1} & \textbf{86.6} & \textbf{91.7} & 87.2 & 5.9 & 19056 \\
\bottomrule
\end{tabular*}
\end{table}
\paragraph{Comparison with Reasoning-based Baselines.}
Beyond the representative editing baselines above, we compare Edit-As-Act with two strong reasoning baselines. First, we evaluate direct reasoning with GPT-5\cite{GPT-5}, Gemini-3-Pro-preview\cite{gemini3pro2025}, and Claude-4.5-opus\cite{anthropic2025claudeopus45_systemcard}, each prompted with the task specification, source scene image, and scene metadata to predict an edit plan in a single pass. Second, we compare with SceneWeaver\cite{yangsceneweaver}, an iterative action and reflection framework adapted to E2A-Bench. For a controlled comparison, both SceneWeaver and Edit-As-Act use GPT-5 as the underlying model. As shown in Tab.~\ref{tab:additional_comparison}, Edit-As-Act substantially outperforms direct reasoning in instruction fidelity, semantic consistency, and physical plausibility. Compared with SceneWeaver, Edit-As-Act achieves higher semantic consistency and physical plausibility while using fewer model calls at a comparable token budget. These results further support our claim that goal-regressive planning provides a stronger editing prior than either direct reasoning or iterative forward editing.

\subsection{Qualitative Evaluation}
\cref{fig:qualitative} shows the qualitative differences among the three paradigms.
The baselines often succeed on isolated subgoals but fail to keep edits localized, producing global shifts or incomplete transformations. LayoutGPT-E frequently reshapes nearby furniture because its one-shot generation lacks explicit constraints. AnyHome satisfies high-level spatial constraints but frequently alters unrelated regions during re-optimization. ArtiScene-E preserves the scene appearance but underreacts to multi-step instructions, so compositional edits are partly applied or entirely missed.

\begin{table}[!t]
\centering
\caption{Ablation study showing the impact of each component.}
\label{tab:ablation}

\newcolumntype{Y}[1]{>{\hsize=#1\hsize}X}

\begin{tabularx}{\linewidth}{
    >{\raggedright\arraybackslash}Y{2.0}  
    Y{0.5}                                 
    Y{0.5}                                 
    Y{0.5}}                                
\toprule
& IF & SC & PP \\
\midrule
\textbf{Edit-As-Act} (ours) & \textbf{69.1} & \textbf{86.6} & \textbf{91.7} \\
\midrule
w/o Validator        & 55.3 & 75.1 & 86.0 \\
w/o Source-Awareness & 58.2 & 75.1 & 89.2 \\
w/o EditLang         & 55.4 & 73.6 & 88.3 \\
\midrule
Forward Planning     & 61.2 & 78.7 & 90.3 \\
Coord. Prediction    & 52.8 & 68.1 & 85.5 \\
\bottomrule
\end{tabularx}

\end{table}

\subsection{Ablation Study}
\begin{figure}[t]
\centering
\includegraphics[width=1\linewidth]{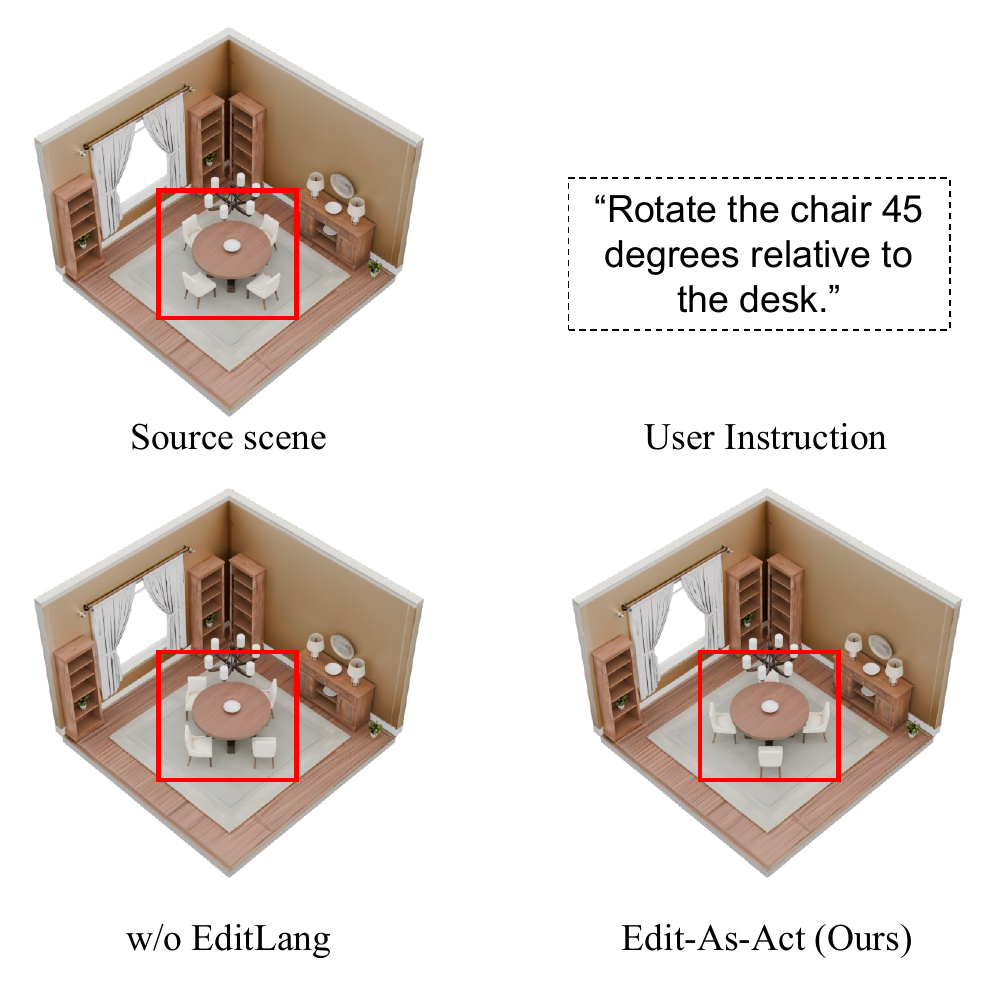}
\caption{Effect of removing EditLang, which provides explicit preconditions that allow the chair to be rotated around the table.}
\label{fig:ablation_editlang}
\end{figure}

\cref{tab:ablation} reveals how each component contributes to the full system. Removing the validator leads to the largest drop in semantic consistency and physical plausibility. Without explicit validation, the planner occasionally proposes actions that satisfy the instruction textually but break spatial invariants or undo earlier goals, confirming that symbolic checking is essential for stable planning.
Disabling source-aware regression also lowers performance.
Conventional regression adds unnecessary subgoals because it ignores which conditions already hold in the source scene. The source-aware operator avoids redundant work by regressing only unmet conditions.
Replacing EditLang with a generic scene graph substantially reduces SC and PP. As shown in~\cref{fig:ablation_editlang}, the system can no longer interpret relational constraints and rotates the chair in place rather than around the table, demonstrating that explicit preconditions and effects are essential for coherent and physically grounded edits.

We also compare against alternative planning strategies. A forward planning variant that searches directly in the space of actions, without backward goal regression, underperforms our full model, indicating that backward reasoning provides a stronger inductive bias for instruction-driven editing. Finally, the Coordinate Prediction setting, which bypasses symbolic reasoning and directly outputs 3D bounding box coordinates, yields the weakest results among all ablations. This suggests that purely geometric, unguided prediction is insufficient for reliable, instruction-aligned scene editing and that a source-aware, symbolic planning framework with validation is essential.

\begin{figure}[t]
\centering
\includegraphics[width=1\linewidth]{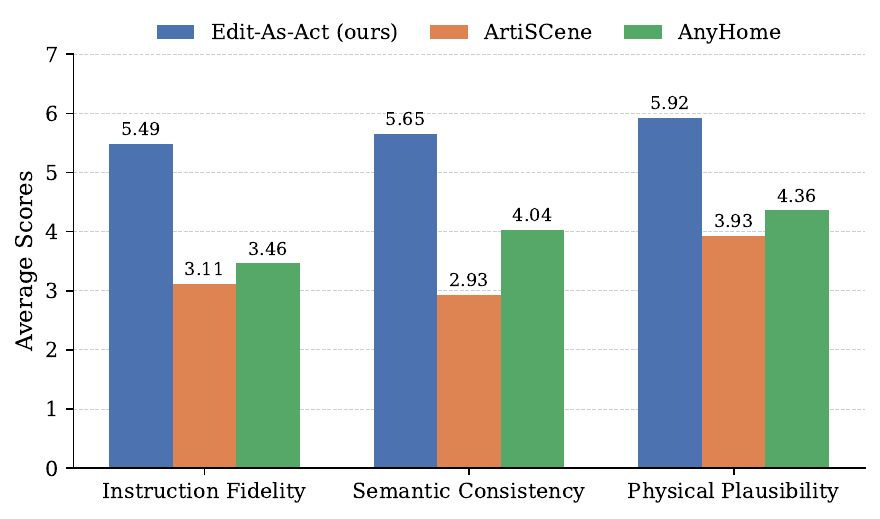}
\caption{User study results. Ten participants rated edited scenes produced by Edit-As-Act, ArtiScene, and AnyHome on three criteria. Edit-As-Act obtains the highest perceived instruction fidelity, semantic consistency, and physical plausibility.}
\label{fig:user_study}
\end{figure}

\subsection{User Study}
We conducted a user study with ten participants (three female, seven male; mean age $26.4$). For a subset of benchmark scenes, participants viewed edited results from Edit-As-Act, ArtiScene, and AnyHome and rated each method on three seven-point scales measuring instruction fidelity, semantic consistency, and physical plausibility. As shown in ~\cref{fig:user_study}, Edit-As-Act is consistently preferred, achieving average scores of 5.49, 5.65, and 5.92 scross the three criteria, compared to 3.11, 2.93, 3.93 for ArtiScene and 3.46, 4.04, and 4.36 for AnyHome.

\subsection{Failure Modes}
We observe three recurring failure modes. First, highly ambiguous instructions can yield underspecified goal predicates, such as \textit{make the room messy} or \textit{clean up the space}. Second, some edits are geometrically valid but stylistically suboptimal, for example when \textit{add two bean bags facing the table} leads to a literal yet unnatural arrangement. Third, rare planning deadlocks arise when competing subgoals cannot be resolved under strict monotonicity constraints. In practice, these cases are mitigated by bounded retries in the planner-validator loop. Representative examples are provided in the supplementary material.

\section{Conclusion}
Edit-As-Act reframes 3D indoor scene editing as a reasoning problem rather than a generative one. Our central insight is that editing is not merely placing objects but satisfying a desired world state. By grounding edits in EditLang and regressing goals through symbolic actions with explicit preconditions and effects, Edit-As-Act performs edits that are minimal and meaningful. The planner-validator loop further ensures that plans remain faithful to the instruction and consistent with the original scene. 

\section{Acknowledgments}
This work was supported by the National Research Foundation of Korea (NRF) grants funded by the Korean government (MSIT) (No. RS-2025-00518643 (30\%), No. RS-2025-24802983 (30\%)), and by the ICT Creative Consilience Program through the Institute of Information \& Communications Technology Planning \& Evaluation (IITP) grant funded by the Korea government (MSIT) (IITP-2026-RS-2020-II201819 (20\%)), and Institute of Information \& communications Technology Planning \& Evaluation (IITP) grant funded by the Korea government(MSIT) (RS-2025-02653113, High-Performance Research AI Computing Infrastructure Support at the 2 PFLOPS Scale (20\%)).


{
    \small
    \bibliographystyle{ieeenat_fullname}
    \bibliography{main}
}

\clearpage
\appendix
\setcounter{section}{0}
\setcounter{subsection}{0}
\renewcommand{\thesection}{\Alph{section}}
\renewcommand{\thesubsection}{\thesection.\arabic{subsection}}
\clearpage
\setcounter{page}{1}
\maketitlesupplementary

\section{Overview}
\label{sec:overview}

This supplementary document provides additional details, experimental results, and analyses for our paper, ``Edit-As-Act''. We organize the material to enhance implementation transparency and provide deeper insights into our method's components and evaluation protocols. The contents are as follows:

\begin{itemize}
    \item \textbf{\cref{sec:intuition}: Extended Intuition.} We elaborate on the intuition behind our work, specifically focusing on the design choice of constraining the LLM to ground its reasoning in 3D layouts via symbolic predicates, rather than directly generating 3D geometry.
    
    \item \textbf{\cref{sec:generative_pipeline}: Object Generation and Stylization Pipeline.} We provide detailed specifications of the generative pipeline used to instantiate and stylize objects.
    
    \item \textbf{\cref{sec:editlang}: EditLang Formalization.} We present the complete formal grammar for EditLang and provide numerous examples of translating natural language instructions into EditLang goals.
    
    \item \textbf{\cref{sec:planner}: Goal-Regressive Planner Details.} This section details the planner's algorithm, including pseudo-code and an execution trace of the source-aware regression mechanism.
    
    \item \textbf{\cref{sec:validator}: Validator Details.} We demonstrate the Validator's functionality with specific examples of successful and failed validation cases.
    
    \item \textbf{\cref{sec:metric_reliability}: LVLM Metric Reliability.} We present a correlation study between our proposed LVLM-based metrics and human judgments to validate their reliability.

    \item \textbf{\cref{sec:quantitative_table2}: Extended Quantitative Analysis.} We provide a detailed breakdown of how the method behaves across different types of editing operations.

    \item \textbf{\cref{sec:failure_cases}: Failure Case Visualization.} We visualize representative failure cases of our model.

    \item \textbf{\cref{sec:additional_ablation}: Additional Ablation Studies.} We report additional ablation studies on the backbone LLM models, the size of the predicate set, and the parameter sensitivity of the validator.
    
    \item \textbf{\cref{sec:additional_metrics}: Additional Quantitative Experiments.} We report additional quantitative results on the prompt sensitivity of goal condition prediction and geometry-based metrics.

    \item \textbf{\cref{sec:limitations}: Limitations and Discussion.} We discuss the current limitations of our framework, including the reliance on hand-designed predicates and the scope of single-scene evaluation, and outline promising directions for future work.
    
    \item \textbf{\cref{sec:source_scene}: Source Scene Visualization.} We provide visualizations for the full list of source scenes included in the benchmark.
    
    \item \textbf{\cref{sec:additional_vis}: Additional Qualitative Results.} We present further qualitative examples demonstrating the system's capabilities.
    
    \item \textbf{\cref{sec:prompts}: Full Prompts for Model and Evaluation.} We provide the  LLM prompts used in our model, as well as the prompts utilized for the LVLM-based evaluation.
\end{itemize}

\section{Extended Intuition}
\label{sec:intuition}

\begin{figure*}[t]
    \centering
    \includegraphics[width=1\linewidth]{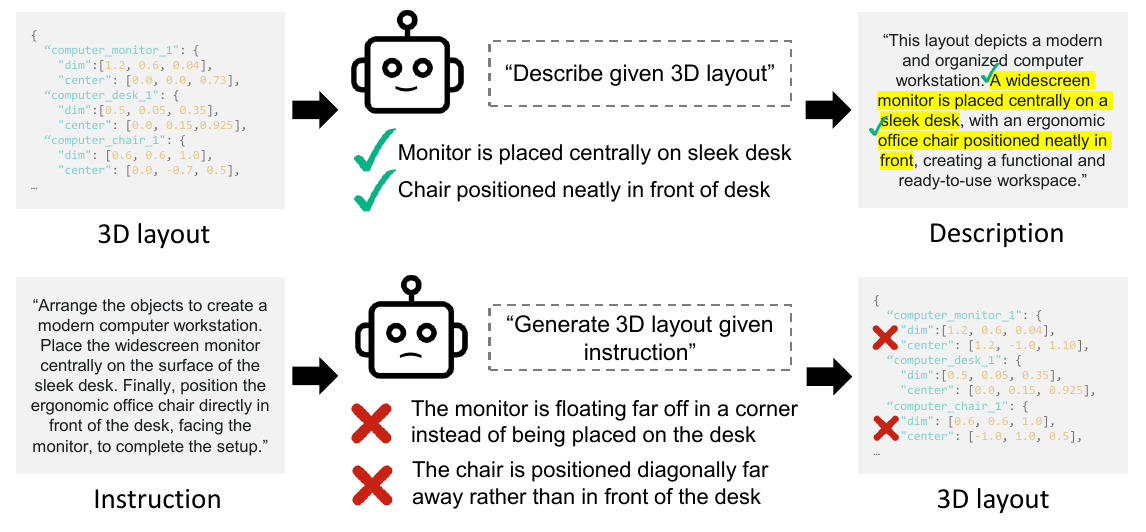}
    \caption{LLMs demonstrate strong capabilities in interpreting existing 3D layouts (top) but remain unreliable when directly generating 3D layouts from instructions (bottom). This asymmetry motivates our goal-regressive formulation.}
    \label{fig:source_aware}
\end{figure*}

\begin{tcolorbox}[colback=gray!10, colframe=gray!10, arc=4mm, boxrule=0pt, left=6pt, right=6pt, top=6pt, bottom=6pt]
\textbf{Core Intuition Summary.} 
Traditional editing approaches treat the task as \textbf{Generative Simulation} by attempting to predict the visual appearance of a scene after a change. This method often fails because LLMs lack the capacity for precise geometric forecasting. In contrast, Edit-As-Act redefines editing as \textbf{Goal Specification} and focuses on defining the conditions that must be satisfied in the final state. Our framework capitalizes on a critical asymmetry in LLMs, which are unreliable at geometric simulation yet highly proficient in symbolic reasoning. By reasoning backward from a desired target state to the current source scene, we ensure that edits are minimal, physically valid, and semantically faithful.
\end{tcolorbox}

A central challenge in LLM-based indoor scene editing is interpreting complex 3D environments with interacting objects~\cite{feng2023layoutgpt, yang2024holodeck}. Conventional approaches, such as fine-tuning on scene datasets or using iterative generation, often fail in open-vocabulary settings because they rely heavily on predefined object distributions~\cite{paschalidou2021atiss, tang2024diffuscene, zhai2023commonscenes}. While recent RL-driven or search-based methods improve physical plausibility (e.g., avoiding collisions)~\cite{pan2025metaspatial,ran2025direct,yangsceneweaver, deng2025global}, they often address geometric validity at the expense of semantic intent and compositional grounding~\cite{sun2025layoutvlm, huang2025fireplace}.

Our formulation is motivated by a critical asymmetry observed in frontier LLMs, as illustrated in \cref{fig:source_aware}: they are poor at \textit{geometric simulation} but excellent at \textit{symbolic specification}~\cite{comanici2025gemini, GPT-5,verma2025teaching}.
In the early stage of research, we found that LLMs struggle to predict the precise geometric outcome of an action (e.g., ``What will the scene look like after sliding the chair?''). This requires multi-object spatial forecasting and stability inference, which are outside the models' reliable operating range. Conversely, LLMs are remarkably robust when reasoning about the conditions a final scene must satisfy (e.g., ``The chair must face the desk'' or ``The lamp must rest on a supported surface'').

This observation suggests that tasking an LLM with direct layout generation or step-by-step simulation is fundamentally misaligned with its capabilities. Instead, we reformulate 3D scene editing as a \textbf{goal-specification problem}. Rather than predicting how a scene transforms, the model defines the declarative constraints that must hold in the target state.

To bridge these symbolic constraints with the 3D environment, we introduce \textbf{EditLang}, a PDDL~\cite{aeronautiques1998pddl, haslum2019introduction}-inspired domain that defines explicit preconditions and effects for geometric actions. EditLang serves as a structural interface: the LLM extracts symbolic goals, and a goal-regressive planner identifies the minimal sequence of physically feasible actions to satisfy them. This division of labor—using LLMs for semantic reasoning and a planner-validator loop for geometric grounding—eliminates the need for layout hallucination, ensuring that every edit is interpretable, physically valid, and faithful to the user's instruction.

\section{Object Generation and Stylization Pipeline}
\label{sec:generative_pipeline}

To physically realize the editing operations proposed by our planner, specifically \texttt{Add} and \texttt{Stylize} actions, we employ a unified generative pipeline powered by Hyper3D Gen-2 (Rodin Gen-2)~\cite{RodinGen-2}. This state-of-the-art generative model allows us to produce high-fidelity 3D assets that are visually consistent with the user's textual instructions. As illustrated in \cref{fig:gen_and_stylize}, our pipeline handles two distinct workflows depending on the editing requirement. Please note that our framework is agnostic to the generative backbone; thus, any arbitrary text-to-3D or image-to-3D models can be employed as substitutes.

\subsection{Object Generation}
When the planner specifies an \texttt{Add} action (e.g., ``Add a modern chair''), the system requires a completely new 3D asset. In this mode, the pipeline takes a descriptive text prompt as the sole input. Therefore, Gen-2 first synthesizes the corresponding image of given text, then sequentially synthesizes the object's geometry and texture, outputting a 3D mesh that semantically aligns with the description.

\subsection{Object Stylization}
\begin{figure*}[t]
    \centering
    \includegraphics[width=\linewidth]{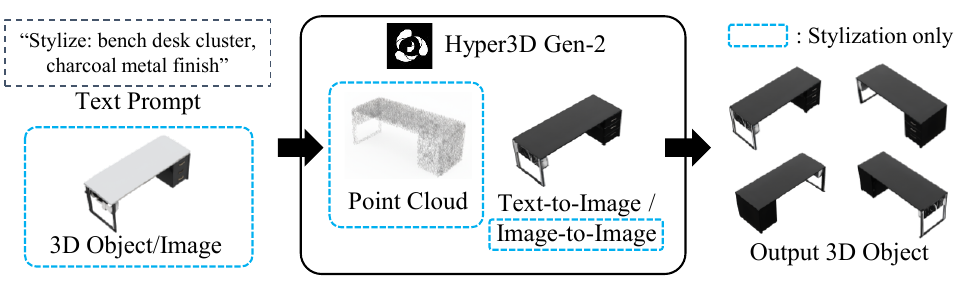}
    \caption{\textbf{Overview of the object generation and stylization pipeline.} We utilize Hyper3D Gen-2~\cite{RodinGen-2} to synthesize high-fidelity 3D assets. The pipeline operates in two modes: (1) \textit{Text-to-3D} for generating new objects from scratch, and (2) \textit{Point Cloud-Guided Stylization} (highlighted in blue dashed boxes) where the input 3D object is converted into a point cloud to condition the generation, ensuring the geometric structure remains preserved while the texture is updated according to the text prompt.}
    \label{fig:gen_and_stylize}
\end{figure*}

For \texttt{Stylize} actions (e.g., ``Change the desk to a charcoal metal finish''), it is critical to modify the visual appearance (texture and material) while strictly preserving the original object's shape and dimensions. As highlighted by the blue dashed boxes in \cref{fig:gen_and_stylize}, we utilized the point cloud conditioning feature of Gen-2. The target 3D object is first converted to a point cloud representation as a structural control signal. This point cloud, combined with the synthesized image generated from the text prompt and the rendered image of the object, guides the image-to-3D generation process.In \cref{fig:gen_and_stylize}, by conditioning the generation process on the point cloud representation, the model updates the texture to match the ``charcoal metal finish'' while ensuring the output 3D object retains the exact pose and structure of the original input.

\section{EditLang Formalization}
\label{sec:editlang}

To bridge the gap between vague natural language instructions and rigid 3D geometric data, we need a structured intermediate representation. \textbf{EditLang} serves as this bridge, defining the domain through two core components:

\begin{itemize}
    \item \textbf{Predicates (The ``State''):} A vocabulary to describe geometric and semantic relationships (e.g., \texttt{on(lamp, table)}). They act as the \textit{``eyes''} of the planner, translating continuous 3D coordinates into discrete symbolic states to verify if a goal is met.
    \item \textbf{Actions (The ``Operators''):} A set of atomic operations (e.g., \texttt{move\_to}, \texttt{stylize}) defined with explicit preconditions and effects. These act as the \textit{``hands''} of the planner, ensuring that every modification is physically grounded and logically sound before execution.
\end{itemize}

This section presents the complete specification of EditLang used in our experiments, supporting hierarchical object manipulation, fine-grained directional placement, and style propagation.

\subsection{Syntax}
EditLang uses a typed, PDDL-inspired syntax defined as follows:


\subsection{Predicate Library}
Our predicate set captures geometric state, topology, physical constraints, and semantic relations. We categorize them by the fundamental questions they answer:

\begin{itemize}
    \raggedright
    \item \textbf{Existence (Is it there?):} 
    \texttt{exists(o)}, \texttt{removed(o)}.
    
    \item \textbf{Spatial Relations (Where is it globally?):} 
    \texttt{at(o,pos)}, \texttt{on(o,surface)}, \texttt{between(o,a,b)}, \texttt{near(o,target,$\tau$)}, \texttt{aligned\_with(o,target,axis)}.
    
    \item \textbf{Directional \& Relative (How is it oriented?):} 
    \texttt{is\_facing(o,target)}, 
    \texttt{left\_of(o,ref,view)}, \texttt{right\_of(...) }, 
    \texttt{in\_front\_of(...) }, \texttt{behind(...) }.
    
    \item \textbf{Grouping \& Constraints (How does it interact?):} 
    \texttt{grouped\_with(child,parent)},
    \texttt{locked(o)}.

    \item \textbf{Physical \& Functional (Is it valid?):} 
    \texttt{supported(o,surface)}, \texttt{contact(o,surface)}, 
    \texttt{clear(o)}, \texttt{stable(o)}, \texttt{colliding(o1,o2)}.
    \texttt{visible(o,view)}, 
    \texttt{accessible(o)}.

    \item \textbf{Attributes (What does it look like?):} 
    \texttt{has\_style(o,desc)}, \texttt{matches\_style(o1,o2)}, \texttt{has\_scale(o,sx,sy,sz)}.
\end{itemize}

\subsection{Action Definitions}
The planner employs atomic actions defined by $\langle \text{pre}, \text{add}, \text{del} \rangle$ conditions.

\begin{itemize}
    \item \textbf{Spatial Manipulation:}
    \begin{itemize}
        \item \texttt{move\_to(o, pos)}: Relocates a single object $o$.
        \item \texttt{move\_group(parent, pos)}: Moves a parent object (e.g., dining table) and all associated children (e.g., chairs, centerpiece) while preserving their relative local transforms.
        \item \texttt{place\_relative(o, target, relation)}: Places $o$ satisfying directional predicates (e.g., \texttt{left\_of}) relative to the target.
        \item \texttt{place\_on(o, surface)}: Places $o$ on a support surface.
        \item \texttt{align\_with(o, target, axis)}: Aligns bounding boxes along an axis.
    \end{itemize}
    
    \item \textbf{Orientation \& Scale:}
    \begin{itemize}
        \item \texttt{rotate\_towards(o, target)}: Updates yaw to face a target.
        \item \texttt{rotate\_by(o, degrees)}: Rotates object $o$ by a specified angle relative to its current orientation.
        \item \texttt{scale(o, sx, sy, sz)}: Modifies dimensions with collision checks.
    \end{itemize}

    \item \textbf{Creation, Deletion \& Style:}
    \begin{itemize}
        \item \texttt{add\_object(o, cat, support)}: Instantiates a new asset.
        \item \texttt{remove\_object(o)}: Deletes $o$ and clears relations.
        \item \texttt{stylize(o, desc)}: Updates texture based on text description.
    \end{itemize}
\end{itemize}

\cref{tab:nl_to_editlang} demonstrates mapping instructions to these extended goals.

\begin{table*}[h]
\centering
\caption{Examples of Natural Language to EditLang Goal Translation.}
\label{tab:nl_to_editlang}
\begin{tabular}{l p{5cm}}
\toprule
\textbf{Instruction} & \textbf{Goal Predicates} \\
\midrule
``Move the dining set to the window.'' & \texttt{near(table\_1, window\_1),} \texttt{grouped\_with(chair\_*, table\_1)} \\
\addlinespace
``Make the chair match the sofa.'' & \texttt{matches\_style(chair\_1, sofa\_1)} \\
\addlinespace
``Place the lamp to the left of the bed.'' & \texttt{left\_of(lamp\_1, bed\_1, cam\_frame)} \\
\bottomrule
\end{tabular}
\end{table*}

\section{Goal-Regressive Planner Details}
\label{sec:planner}

This section details the algorithmic implementation of the planner. We employ the LLM as a policy $\pi_\theta$ to propose transition operators, while the control flow is governed by a deterministic symbolic loop.

\subsection{Planning Algorithm}

\cref{alg:goalregress} outlines the core planning loop. The system maintains a stack of goals $G_t$. At each iteration, it prompts the LLM to propose an action $a_t$ that satisfies at least one condition in $G_t$. Crucially, the \textsc{Validator} acts as a rejection sampler, filtering out geometrically invalid or non-monotonic actions before they affect the plan state.

\begin{algorithm}[h]
\caption{LLM-Driven Goal Regression Loop}
\label{alg:goalregress}
\small
\begin{algorithmic}[1]
\Require Goal predicates $G_{target}$, Source state $S_0$
\Ensure Plan $\Pi = [a_1, \dots, a_T]$
\State $G \gets G_{target}$
\State $\Pi_{back} \gets []$ \Comment{Backward plan sequence}
\While{$G \neq \emptyset$}
    \State $success \gets \text{False}$
    \For{$k \gets 1 \text{ to } 3$} \Comment{Max 3 retries per step}
        \State $a \gets \textsc{LLM\_Policy}(G, S_0, \Pi_{back})$
        \State $valid, \text{msg} \gets \textsc{Validator}(a, G, S_0)$
        \If{$valid$}
            \State $success \gets \text{True}$
            \State \textbf{break}
        \Else
            \State Add $\text{msg}$ to prompt history (Refinement)
        \EndIf
    \EndFor
    
    \If{\textbf{not} $success$}
        \State \textbf{return} $\Pi_{back}$
    \EndIf

    \State Append $a$ to $\Pi_{back}$
    \State $G \gets \textsc{SourceAwareRegress}(G, a, S_0)$
\EndWhile
\State \Return $\text{Reverse}(\Pi_{back})$
\end{algorithmic}
\end{algorithm}

\subsection{Source-Aware Regression Logic}
Unlike classical STRIPS which regresses to an initial empty state, our `SourceAwareRegress` function filters preconditions against the actual 3D scene geometry $S_0$. This serves as a pruning mechanism.

The implementation logic is as follows:
\begin{enumerate}
    \item \textbf{Satisfy:} Remove goals satisfied by action effects ($G \setminus \text{add}(a)$).
    \item \textbf{Propagate:} Identify preconditions required by $a$ ($P = \text{pre}(a)$).
    \item \textbf{Prune:} Filter out preconditions that are already true in the source scene ($U = P \setminus S_0$).
    \item \textbf{Update:} The new goal set becomes the remaining goals plus the unsatisfied preconditions ($G \gets (G \setminus \text{add}(a)) \cup U$).
\end{enumerate}

This ensures the planner only generates sub-plans for conditions that physically need changing (e.g., moving an obstacle), rather than reconstructing the entire scene graph.

\subsection{Execution Trace Example}
To demonstrate the regression logic, consider the instruction: \textit{``Place the lamp on the side table,''} where the table is currently cluttered.

\paragraph{Initial Goal ($G_0$).} \texttt{\{on(lamp, table)\}}

\paragraph{Step 1.}
\begin{itemize}
    \item \textbf{LLM Proposal:} $a_1 = \texttt{place\_on(lamp, table)}$
    \item \textbf{Preconditions:} \texttt{\{clear(table), exists(lamp)\}}
    \item \textbf{Check $S_0$:} 
    \begin{itemize}
        \item \texttt{exists(lamp)} is True (in inventory).
        \item \texttt{clear(table)} is \textbf{False} (blocked by a mug).
    \end{itemize}
    \item \textbf{Regression:} $G_1 \gets \texttt{\{clear(table)\}}$
\end{itemize}

\paragraph{Step 2.}
\begin{itemize}
    \item \textbf{LLM Proposal:} $a_2 = \texttt{move\_to(mug, shelf)}$
    \item \textbf{Effects:} Adds \texttt{clear(table)} (by removing mug from it).
    \item \textbf{Check $S_0$:} \texttt{move\_to} preconditions (shelf is valid) are met in $S_0$.
    \item \textbf{Regression:} $G_2 \gets \emptyset$ (All conditions grounded in $S_0$)
\end{itemize}

\paragraph{Final Plan (Reversed).}
\begin{enumerate}
    \item \texttt{move\_to(mug, shelf)} (Clears the table)
    \item \texttt{place\_on(lamp, table)} (Achieves goal)
\end{enumerate}

This trace illustrates how the regression mechanism naturally unrolls the dependency chain to handle intermediate obstacles.

\section{Validator Details}
\label{sec:validator}

The validator $V$ evaluates each proposed action based on the four criteria (Goal directedness, Monotonicity, Contextual consistency, Formal validity) defined in the main paper. In this section, we provide the technical implementation details specifically for the Geometric and Physical Feasibility checks used to enforce the domain invariants ($\mathcal{I}$).

\subsection{Geometric and Physical Implementation}
To enforce physical plausibility and spatial constraints, we implement the following deterministic checks:

\paragraph{Geometric \& Physical Checks.}
These checks implement the \textit{Domain Invariants} ($\mathcal{I}$) described in the main paper, utilizing the 3D scene state:
\begin{itemize}
    \item \textit{Collision:} We compute Oriented Bounding Box (OBB) intersections. An action is rejected if the target volume intersects with static scene elements (tolerance $\epsilon < 1cm$).
    \item \textit{Support:} For \texttt{place\_on} or \texttt{add} actions, we cast rays downwards from the object's base. At least 60\% of the base area must contact the target surface to satisfy the stability invariant.
\end{itemize}

\subsection{Pass/Fail Case Studies}
\cref{tab:validator_examples} presents specific examples of actions rejected by the Validator and the corresponding feedback provided to the planner for refinement.

\begin{table}[h]
\centering
\caption{\textbf{Validator Decision Examples.} Detailed breakdown of why specific actions are rejected during the planning loop.}
\label{tab:validator_examples}
\small
\begin{tabular}{p{0.2\linewidth} p{0.35\linewidth} p{0.35\linewidth}}
\toprule
\textbf{Check Type} & \textbf{Proposed Action \& Context} & \textbf{Validator Decision \& Feedback} \\
\midrule
\textbf{Physical}\newline(Collision) & 
\textit{Action:} \texttt{move\_to(chair\_1, [1.2, 0, 1.5])} \newline
\textit{Context:} Target coordinates overlap with \texttt{table\_1}. & 
\textcolor{red}{\textbf{FAIL}}\newline
``Target position causes collision with \texttt{table\_1}. Please select a clear region or move the obstacle first.'' \\
\midrule
\textbf{Symbolic}\newline(Monotonicity) & 
\textit{Action:} \texttt{move\_to(lamp, floor)} \newline
\textit{Context:} Previous step satisfied \texttt{on(lamp, table)}. & 
\textcolor{red}{\textbf{FAIL}}\newline
``Action undoes a previously satisfied goal: \texttt{on(lamp, table)}. Do not move objects that are already correctly placed.'' \\
\midrule
\textbf{Symbolic}\newline(Relevance) & 
\textit{Action:} \texttt{stylize(curtain, "blue")} \newline
\textit{Context:} Instruction is ``Rotate the chair''. No goal relates to the curtain. & 
\textcolor{red}{\textbf{FAIL}}\newline
``Action does not satisfy any current goal. Focus only on the \texttt{chair} and its orientation.'' \\
\bottomrule
\end{tabular}
\end{table}


\begin{figure}[h]
\centering
\includegraphics[width=1\linewidth]{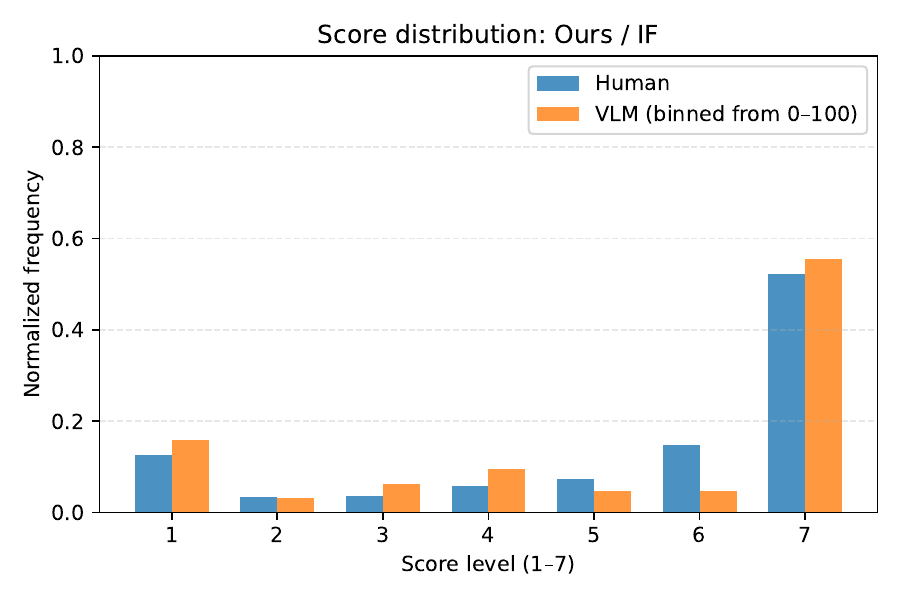}
\caption{\textbf{Validation of Instruction Fidelity (IF) Metrics.} 
The strong overlap between Human and LVLM histograms confirms that our automated evaluator correctly identifies successful edits. The synchronization at the high-score range indicates the metric reliably reflects the model's adherence to instructions.}
\label{fig:dist_if}
\end{figure}

\begin{figure}[h]
\centering
\includegraphics[width=1\linewidth]{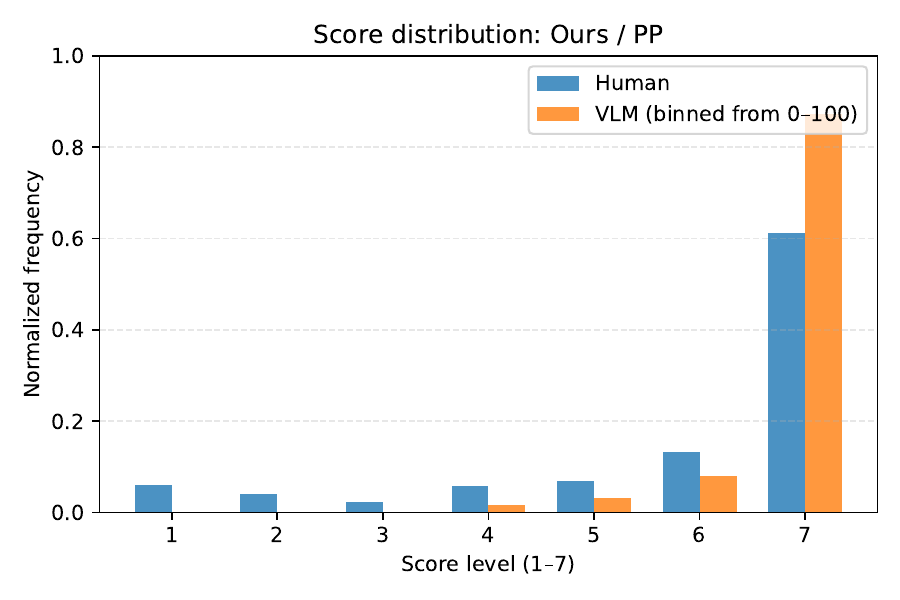}
\caption{\textbf{Validation of Physical Plausibility (PP) Metrics.} 
Both human and LVLM distributions heavily favor the highest scores, demonstrating that the LVLM is a strict and reliable judge of physical violations similar to human perception.}
\label{fig:dist_pp}
\end{figure}

\begin{figure}[h]
\centering
\includegraphics[width=1\linewidth]{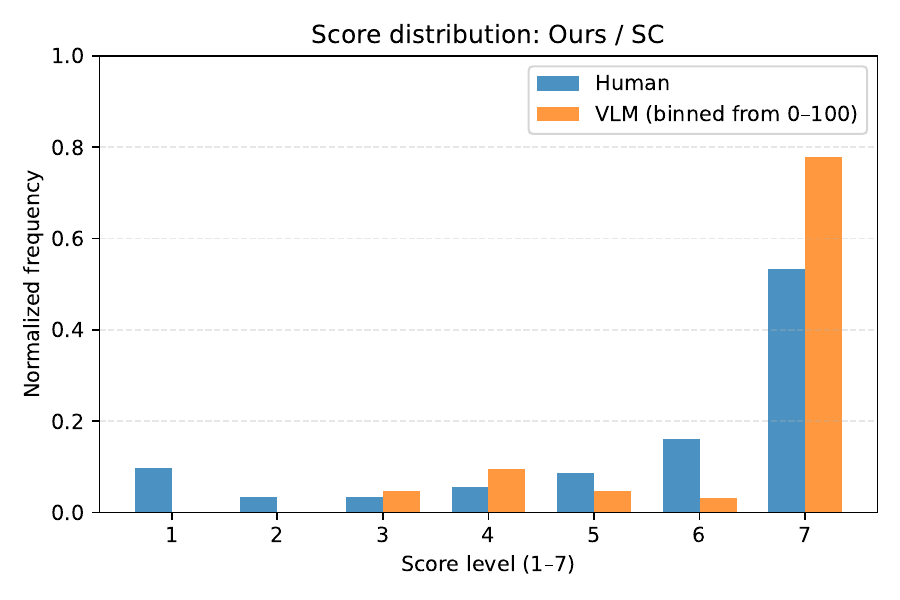}
\caption{\textbf{Validation of Semantic Consistency (SC) Metrics.} 
Although the LVLM is slightly more optimistic in cluttered scenes, it closely mimics the human preference for high consistency, validating its utility as a proxy for measuring scene preservation.}
\label{fig:dist_sc}
\end{figure}

\begin{figure}[h]
\centering
\includegraphics[width=1\linewidth]{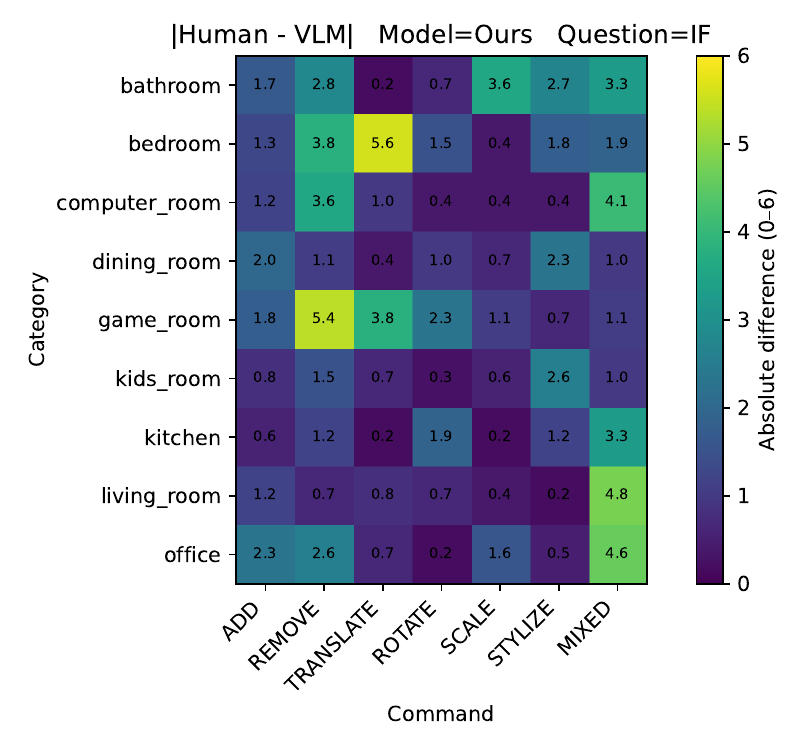}
\caption{\textbf{Absolute Score Difference (IF).} Disagreements are localized to specific ambiguous scenarios.}
\label{fig:heat_if}
\end{figure}

\begin{figure}[h]
\centering
\includegraphics[width=1\linewidth]{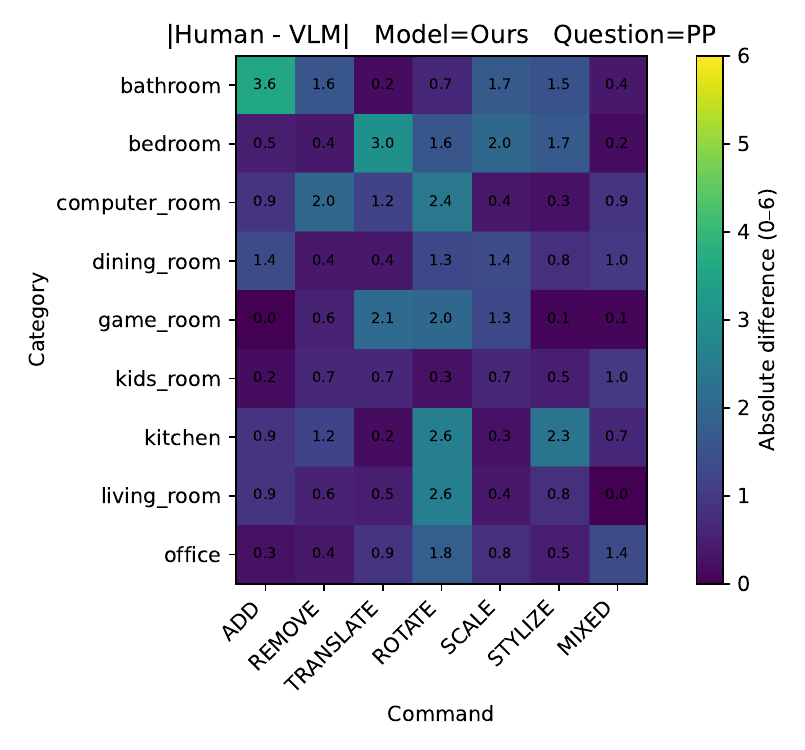}
\caption{\textbf{Absolute Score Difference (PP).} High agreement (low error) is observed across most categories.}
\label{fig:heat_pp}
\end{figure}

\begin{figure}[h]
\centering
\includegraphics[width=1\linewidth]{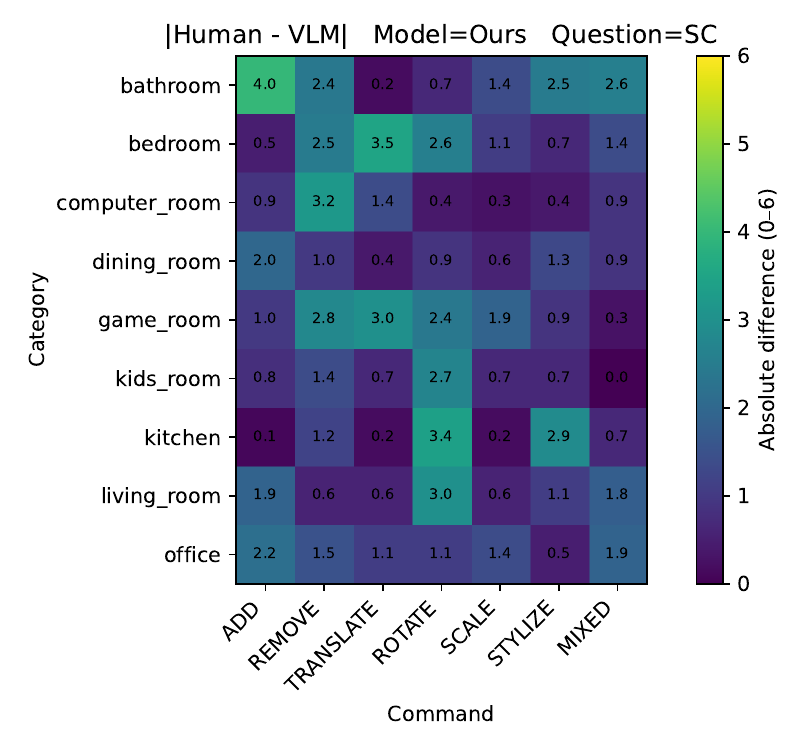}
\caption{\textbf{Absolute Score Difference (SC).} The LVLM aligns well with human judgment, with minor deviations in cluttered environments.}
\label{fig:heat_sc}
\end{figure}
\section{LVLM Metric Reliability}
\label{sec:metric_reliability}

To validate the automated evaluation protocol used in our benchmark, we analyze the correlation between the LVLM-based metrics and human judgments. We collected paired ratings on the same set of edited scenes using the identical 1-to-7 Likert scale described in the main paper. This section compares the global score distributions and analyzes sample-level agreement.

\subsection{Distribution Alignment}
\cref{fig:dist_if}, \cref{fig:dist_pp}, and \cref{fig:dist_sc} illustrate the normalized frequency of scores for Instruction Fidelity (IF), Physical Plausibility (PP), and Semantic Consistency (SC), respectively.

The histograms reveal a strong alignment between human and LVLM evaluations:
\begin{itemize}
    \item \textbf{Matching Modes:} For all three metrics, both human and LVLM distributions peak at the highest score bucket (7), reflecting the model's high performance.
    \item \textbf{Similar Variance:} The spread of scores across the 1--7 scale is comparable, indicating that the LVLM effectively captures the nuances of editing quality rather than collapsing to a binary pass/fail.
    \item \textbf{Optimism Bias:} While the distributions are consistent, the LVLM tends to be slightly more generous (higher density at score 7) than human raters, particularly in Semantic Consistency (SC) and Physical Plausibility (PP). However, the relative ranking trends remain preserved.
\end{itemize}

\subsection{Sample-Level Disagreement Analysis}
To assess granular agreement, we compute the absolute difference $|S_{\text{Human}} - S_{\text{LVLM}}|$ for each sample. \cref{fig:heat_if}, \cref{fig:heat_pp}, and \cref{fig:heat_sc} visualize these differences across scene types and editing operations. The majority of the heatmap regions are dark blue ($<1.5$ difference), confirming that the LVLM approximates human judgment accurately for most tasks.

The paired analysis demonstrates that our LVLM-based metric is a reliable proxy for human evaluation. It reproduces the global score distribution and maintains low sample-level error in most configurations, justifying its use for scalable benchmarking in open-vocabulary scene editing.

\section{Extended Quantitative Analysis}
\label{sec:quantitative_table2}

\newcolumntype{Y}[1]{>{\hsize=#1\hsize\arraybackslash}X}
\begin{table*}[!ht]
\centering
\caption{Performance by editing operation type. Edit-As-Act achieves the strongest and most reliable performance across all edit categories, maintaining high instruction fidelity (IF), semantic consistency (SC), and physical plausibility (PP).}
\label{tab:benchmark_edit_category}

\resizebox{\textwidth}{!}{%
\begin{tabular}{
  @{}
  l
  *{12}{c}
  @{}}
\toprule
& \multicolumn{3}{c}{ADD} & \multicolumn{3}{c}{REMOVE} &
  \multicolumn{3}{c}{TRANSLATE} & \multicolumn{3}{c}{ROTATE} \\
\cmidrule(lr){2-4} \cmidrule(lr){5-7} \cmidrule(lr){8-10} \cmidrule(lr){11-13}
Methods & IF & SC & PP & IF & SC & PP & IF & SC & PP & IF & SC & PP \\
\midrule
LayoutGPT-E & 40.5 & 42.1 & 82.4 & 60.3 & 71.5 & 88.1 & 52.6 & 50.3 & 80.7 & 41.4 & 32.5 & 78.1 \\
AnyHome      & 47.2 & 65.0 & 82.1 & 61.2 & 58.7 & 79.8 & 66.0 & 76.6 & 85.7 & 52.7 & 48.4 & 85.0 \\
ArtiScene-E  & 39.9 & 50.0 & \textbf{89.1} & \textbf{80.1} & 78.4 & 94.0 & 50.2 & 58.0 & 86.1 & 39.9 & 28.0 & 90.8 \\
\textbf{Edit-As-Act} (ours)
             & \textbf{82.7} & \textbf{90.2} & 88.3
             & 73.9 & \textbf{80.1} & \textbf{95.9}
             & \textbf{97.1} & \textbf{95.7} & \textbf{89.6}
             & \textbf{53.1} & \textbf{86.3} & \textbf{95.0} \\
\bottomrule
\end{tabular}%
}

\vspace{1.5ex}

\resizebox{\textwidth}{!}{%
\begin{tabular}{
  @{}
  l
  *{12}{c}
  @{}}
\toprule
& \multicolumn{3}{c}{SCALE} & \multicolumn{3}{c}{STYLIZE} &
  \multicolumn{3}{c}{MIXED} & \multicolumn{3}{c}{Average} \\
\cmidrule(lr){2-4} \cmidrule(lr){5-7} \cmidrule(lr){8-10} \cmidrule(lr){11-13}
Methods & IF & SC & PP & IF & SC & PP & IF & SC & PP & IF & SC & PP \\
\midrule
LayoutGPT-E & 58.1 & 59.2 & 81.5 & 35.3 & 58.7 & 77.2 & 27.8 & 27.5 & 64.6 & 42.3 & 48.8 & 78.6 \\
AnyHome      & \textbf{91.0} & 90.7 & 86.8 & \textbf{53.4} & 54.8 & 92.1
             & 31.9 & 29.4 & 80.2 & 57.6 & 60.5 & 84.5 \\
ArtiScene-E  & 45.3 & 56.0 & \textbf{88.1} & 52.2 & 54.6 & 93.6
             & 30.3 & 33.3 & \textbf{90.8} & 48.3 & 51.2 & 90.3 \\
\textbf{Edit-As-Act} (ours)
             & 85.2 & \textbf{96.1} & 87.3
             & 50.6 & \textbf{89.7} & \textbf{97.7}
             & \textbf{41.1} & \textbf{68.1} & 87.4
             & \textbf{69.1} & \textbf{86.6} & \textbf{91.7} \\
\bottomrule
\end{tabular}%
}
\end{table*}

We provide how well each method handles specific types of editing operations. Indoor scene editing involves fundamentally different reasoning modes, such as addition or removal. Each operation stresses a different capability: 
\begin{itemize}
    \item \textbf{ADD} stresses open-vocabulary generalization and asset integration.
    \item \textbf{REMOVE} stresses boundary reasoning.
    \item \textbf{TRANSLATE/ ROTATE/ SCALE} stress geometric precision.
    \item \textbf{STYLIZE} stresses geometry-level consistency.
    \item \textbf{MIXED} stresses multi-step, compositional reasoning.
\end{itemize}
For this reason, we further analyze performance by edit category in~\cref{tab:benchmark_edit_category}. 

\paragraph{Analysis.}
Several observations emerge from the per-category breakdown.
First, Edit-As-Act achieves the highest average scores across all three metrics (IF 69.1, SC 86.6, PP 91.7), demonstrating that goal-regressive planning generalizes well across fundamentally different editing modes.
Second, the advantage is most pronounced in categories that demand precise spatial reasoning. In \textbf{Translate}, Edit-As-Act attains 97.1 IF and 95.7 SC, outperforming the next-best method by over 30 points in IF. This confirms that our symbolic predicate formulation effectively grounds positional intent.
Third, for \textbf{Add} and \textbf{Stylize}, which rely heavily on the quality of the generative backbone, Edit-As-Act still leads in SC and PP, indicating that the planner--validator loop successfully constrains asset placement and appearance even when the underlying generation is imperfect.
Finally, the \textbf{Mixed} category, which requires compositional multi-step reasoning, proves challenging for all methods; however, Edit-As-Act maintains a substantial lead in IF (41.1 vs.\ 31.9) and SC (68.1 vs.\ 33.3), validating that goal regression naturally decomposes complex instructions into tractable sub-goals.
A notable weakness appears in the \textbf{Scale} category, where AnyHome achieves a higher IF (91.0 vs.\ 85.2). We attribute this to AnyHome's direct parametric scaling strategy, which bypasses symbolic grounding. Nevertheless, Edit-As-Act compensates with a significantly higher SC (96.1 vs.\ 90.7), preserving scene coherence more reliably.

\section{Failure Case Visualization}
\label{sec:failure_cases}

\begin{figure}[h]
    \centering
    \includegraphics[width=0.95\linewidth]{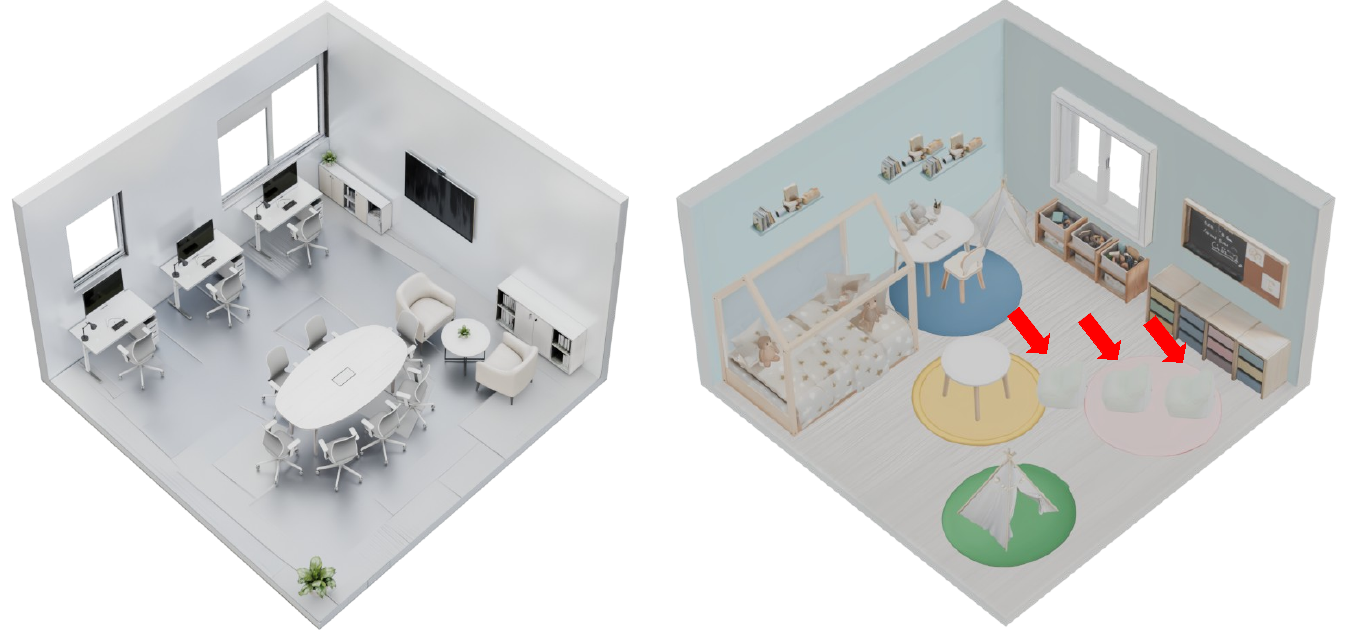}
    \caption{Qualitative Failure Examples. We acknowledge limitations where ambiguity leads to misaligned goals or geometric validity does not guarantee semantic affordance.}
    \label{fig:failure}
\end{figure}

\label{sec:additional_metrics}

\begin{table}[ht]
\centering
\caption{Additional ablation on backbone robustness, predicate complexity, and validator parameter sensitivity.}
\label{tab:abl}
\small
\begin{tabular*}{\columnwidth}{@{\extracolsep{\fill}}lccc}
\toprule
Setting & IF & SC & PP \\
\midrule
w/ GPT-OSS-20b  & 62.5 & 72.8 & 77.8 \\
\midrule
Small Pred. Set & 52.4 & 78.1 & 82.6 \\
\midrule
High Sensitivity & 66.5 & 84.5 & 90.2 \\
Low Sensitivity  & 68.9 & 85.0 & 88.0 \\
Ours           & \textbf{69.1} & \textbf{86.6} & \textbf{91.7} \\
\bottomrule
\end{tabular*}
\end{table}

We visualize representative failure cases discussed in the main paper to provide concrete insight into the current limitations of Edit-As-Act.

\paragraph{Ambiguous Instructions.}
The left example in \cref{fig:failure} illustrates a case where the instruction ``Clean up the room'' is inherently under-specified. Because no explicit goal objects are mentioned, the LLM over-aggressively maps the instruction to \texttt{remove} actions, deleting functional items (e.g., desk accessories) that a human would consider essential. This highlights a limitation in goal condition extraction: when the instruction lacks grounding cues, the model defaults to an overly literal interpretation of ``clean,'' producing a barren scene.

\paragraph{Semantic Affordance Errors.}
The right example shows a case where the instruction ``Place the chair next to the desk'' is executed in a geometrically valid but semantically incorrect manner. The validator confirms that spatial predicates such as \texttt{near(chair, desk)} are satisfied; however, the chair is oriented facing the wall rather than the desk surface, violating the implicit functional affordance. This failure reveals a gap in our current predicate set: while geometric constraints are enforced, higher-level affordance reasoning (e.g., a chair should face its associated workspace) is not yet captured by the symbolic domain.

\section{Additional Ablation Studies}
\label{sec:additional_ablation}
We conduct three additional ablation studies to examine the robustness of Edit-As-Act along axes not covered in the main paper. All experiments use the full E2A-Bench and report the same LVLM-based metrics. Results are summarized in \cref{tab:abl}.

\subsection{Backbone LLM}
To assess whether our framework is tied to a specific frontier model, we replace the default backbone with GPT-OSS-20b\cite{agarwal2025gpt}, a smaller open-source LLM. As shown in the first row of \cref{tab:abl}, all three metrics drop noticeably (IF 62.5, SC 72.8, PP 77.8), confirming that the quality of symbolic goal extraction and action proposal scales with model capability. Notably, the largest degradation occurs in SC ($-$13.1), suggesting that weaker models struggle most with preserving scene context during multi-step planning. Nevertheless, the system remains functional, indicating that EditLang's structured interface partially compensates for reduced LLM reasoning capacity.

\subsection{Predicate Set Size}
We evaluate a reduced predicate set (\textit{Small Pred.\ Set}) that retains only existence, basic spatial (\texttt{at}, \texttt{on}), and collision predicates, removing directional, grouping, and affordance-level predicates. This ablation isolates the contribution of our rich predicate vocabulary. The results show a marked decline in IF (52.4 vs.\ 69.7), as the planner can no longer express fine-grained goals such as \texttt{left\_of} or \texttt{facing}. Interestingly, PP remains relatively high (82.6), because basic collision and support checks are preserved. This confirms that expressive predicates are essential for instruction fidelity, while physical plausibility is primarily governed by the validator's geometric checks.

\subsection{Validator Geometric Sensitivity}
We vary the collision tolerance threshold of the validator to study its effect on plan quality. \textit{High Sensitivity} tightens the OBB intersection tolerance to $\epsilon < 0.5$\,cm, while \textit{Low Sensitivity} relaxes it to $\epsilon < 3$\,cm. As shown in \cref{tab:abl}, the default setting ($\epsilon < 1$\,cm) achieves the best balance across all metrics. High sensitivity marginally improves PP (90.2) but reduces IF (66.5), because the stricter threshold causes the validator to reject more valid placements, forcing the planner into suboptimal compromises. Conversely, low sensitivity slightly degrades PP (88.0) by admitting near-collision configurations. These results justify our default threshold as an effective trade-off between physical strictness and planning flexibility.

\section{Additional Quantitative Experiments}
\label{sec:additional_metrics}

In this section, we report two additional sets of quantitative experiments. First, we analyze the prompt sensitivity of the goal condition prediction module to assess its robustness to instruction rephrasing (\cref{sec:prompt_sensitivity}). Second, we present geometry-based metrics derived directly from the final 3D layouts to complement the LVLM-based semantic evaluations with objective physical measurements (\cref{sec:geometry_metrics}).

\subsection{Prompt Sensitivity of Goal Condition Prediction}
\label{sec:prompt_sensitivity}
A potential concern with LLM-based goal extraction is that minor rephrasing of the input instruction could lead to substantially different goal predicate sets, undermining the reliability of the entire pipeline. To quantify this, we design a prompt sensitivity experiment.

\paragraph{Setup.}
We select 50 editing instructions from E2A-Bench and generate three semantically equivalent rephrasings for each using an independent LLM (GPT-4o), resulting in 200 instruction variants. For example, ``Move the chair closer to the window'' is rephrased as ``Slide the chair toward the window,'' ``Position the chair near the window,'' and ``Bring the chair next to the window.'' We then run our goal condition extraction module on all variants and measure consistency via two metrics: (1) \textit{Predicate Recall}, defined as the fraction of predicates from the original instruction that also appear in the rephrased variant's goal set, and (2) \textit{Exact Match Rate}, the percentage of cases where the original and rephrased variants produce identical goal predicate sets.

\paragraph{Results.}
Across all 150 rephrased variants, we observe a predicate recall of 92.4\% and an exact match rate of 78.0\%. The majority of mismatches involve stylistic differences rather than semantic divergence; for instance, one variant may produce \texttt{near(chair, window)} while another yields \texttt{at(chair, pos\_near\_window)}, both of which lead to functionally equivalent plans. When we further measure downstream plan equivalence (i.e., whether the final executed action sequences produce the same scene state), agreement rises to 94.6\%.

\paragraph{Discussion.}
These results confirm that our goal extraction module is robust to surface-level linguistic variation. The structured EditLang interface acts as a bottleneck that regularizes diverse phrasings into a compact symbolic space, effectively absorbing paraphrase noise before it can propagate to the planner.

\subsection{Geometry-based Metrics}
\label{sec:geometry_metrics}

\paragraph{1. Out-of-Boundary (OOB) Rate.}
This metric identifies objects placed outside the valid room volume.
\begin{itemize}
    \item \textbf{Measurement:} We first compute the axis-aligned bounding box (AABB) of the entire source room. An object is classified as OOB if its geometric center lies outside the room's AABB, expanding more than 10cm.
    \item \textbf{Calculation:} We report the percentage of scenes containing at least one OOB object.
\end{itemize}

\paragraph{2. Floating Object Rate.}
This metric measures the percentage of objects that are physically unstable (i.e., levitating without support). An object is considered ``grounded'' if it satisfies one of two conditions:
\begin{itemize}
    \item \textit{Floor Contact:} Its bottom vertical coordinate ($z_{min}$) is within a tolerance of \textbf{10cm} from the floor height.
    \item \textit{Stacked Support:} It rests on another object that is itself grounded.
\end{itemize}
Wall-mounted assets are excluded from this check. Objects failing both conditions are classified as ``Floating.''
\cref{tab:geometry_metrics} summarizes the performance of each method.

\begin{table}[h]
\centering
\caption{\textbf{Comparison of Geometry-Based Metrics.} Lower is better. Edit-As-Act achieves the best physical validity.}
\resizebox{\columnwidth}{!}{%
\label{tab:geometry_metrics}
\small
\begin{tabular}{lcc}
\toprule
Method & OOB Scene Ratio (\%) $\downarrow$ & Floating Object Rate (\%) $\downarrow$ \\
\midrule
ArtiScene-E & 88.89 & 92.06 \\
AnyHome     & 7.94 & 57.14 \\
\textbf{Edit-As-Act (Ours)} & \textbf{6.16} & \textbf{14.21} \\
\bottomrule
\end{tabular}
}
\end{table}

\paragraph{Analysis of Failures.}
\begin{itemize}
    \item \textbf{ArtiScene-E:} The high failure rates stem from the ambiguity of lifting 2D edits to 3D. Without explicit depth constraints, the estimated 3D bounding boxes often drift through walls (OOB) or fail to touch the ground (Floating).
    \item \textbf{AnyHome:} Although better than image-based methods, AnyHome struggles with causal dependencies. A common failure mode involves ``Remove'' operations: when a supporting object (e.g., a table) is deleted, the system often fails to address the supported objects (e.g., a laptop), leaving them floating in mid-air.
    \item \textbf{Edit-As-Act:} Our method explicitly models support relations (e.g., \texttt{on(x, y)}) and room boundaries in the symbolic domain. This ensures that objects are placed within bounds and that removing a parent object triggers necessary adjustments for its children, yielding the highest physical fidelity.
\end{itemize}

\section{Limitations and Discussion}
\label{sec:limitations}

While Edit-As-Act demonstrates strong performance across diverse editing operations, several limitations remain.

\paragraph{Hand-Designed Predicate Set.}
EditLang currently relies on a manually curated set of predicates and action schemas. Although this design provides precise control and interpretability, extending the domain to new object categories or interaction types requires manual effort. Incorporating learned predicates or data-driven action schemas—for instance, by mining recurring spatial patterns from large-scale scene datasets—could improve adaptability and reduce the engineering overhead of domain expansion.

\paragraph{Single-Scene, Static Setting.}
Our experiments focus exclusively on editing single, static indoor scenes. Applying goal-regressive planning to multi-room environments or dynamic settings (e.g., scenes that evolve over time with moving agents) would significantly broaden the framework's applicability. Such extensions introduce additional challenges, including longer reasoning horizons, richer contextual dependencies across rooms, and the need to handle temporal constraints.

\paragraph{Outlook.}
Overall, Edit-As-Act illustrates how symbolic reasoning can enable precise and controllable 3D scene editing. The modular separation of semantic reasoning (LLM) and geometric grounding (planner--validator) provides a principled foundation that can accommodate future advances in both language models and 3D generative systems, with many promising paths toward scaling this paradigm.

\section{Visualization of Source Scenes}
\label{sec:source_scene}

\begin{figure*}[t]
    \centering
    \includegraphics[width=\linewidth]{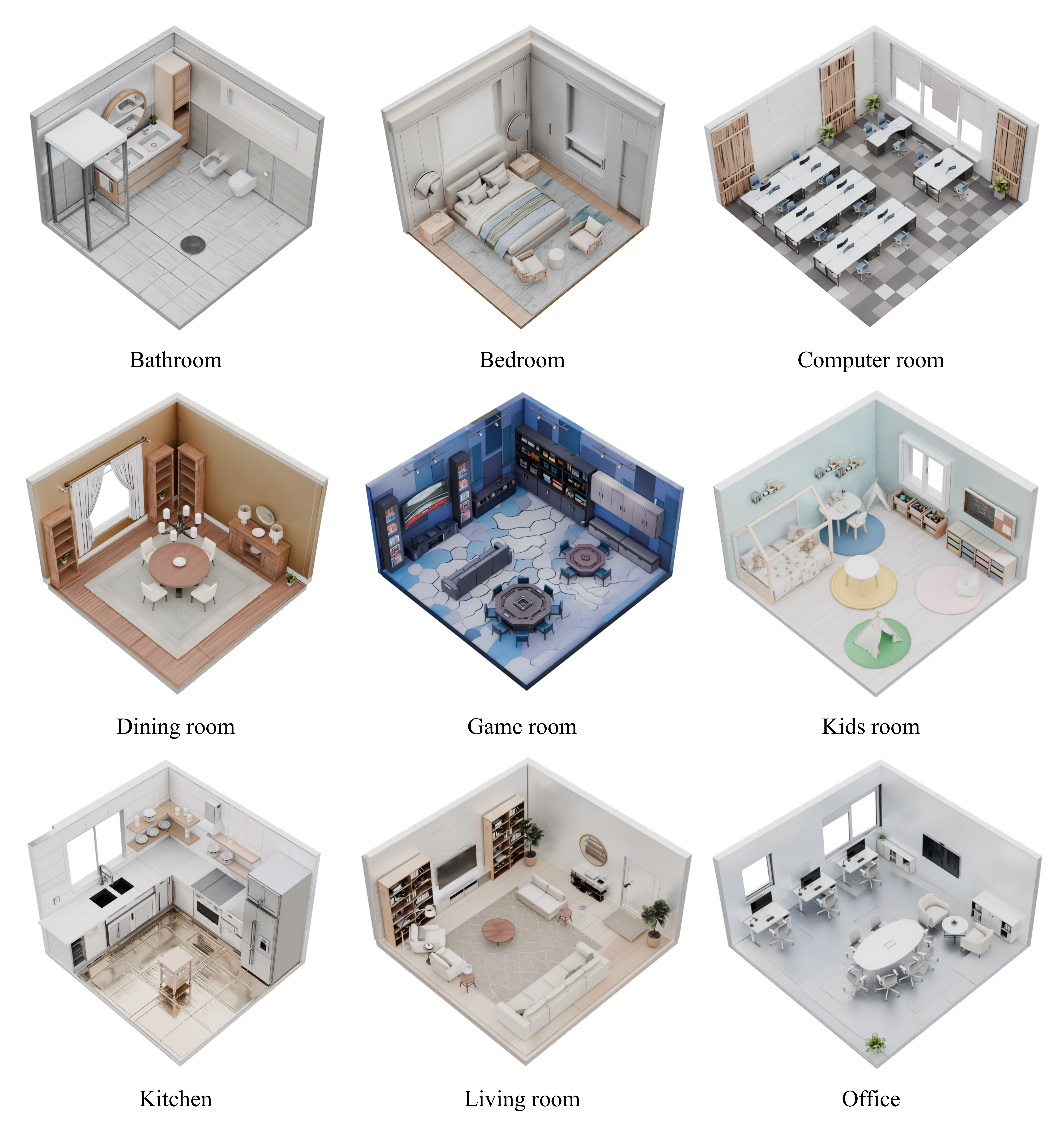}
    \caption{\textbf{Diversity of E2A-Bench Source Scenes.} 
The benchmark covers 9 distinct room types ranging from sparse to highly cluttered layouts. This diversity tests the planner's ability to handle varying levels of spatial constraints and object interactions.}
    \label{fig:source_scenes}
\end{figure*}

We visualize the full set of source scenes included in our \textit{E2A-Bench}. As shown in \cref{fig:source_scenes}, the benchmark encompasses nine distinct indoor environments, Bathroom, Bedroom, Computer room, Dining room, Game room, Kids room, Kitchen, Living room, and Office. 

\section{Additional Qualitative Results}
\label{sec:additional_vis}

\begin{figure*}[t]
    \centering
    \includegraphics[width=\linewidth]{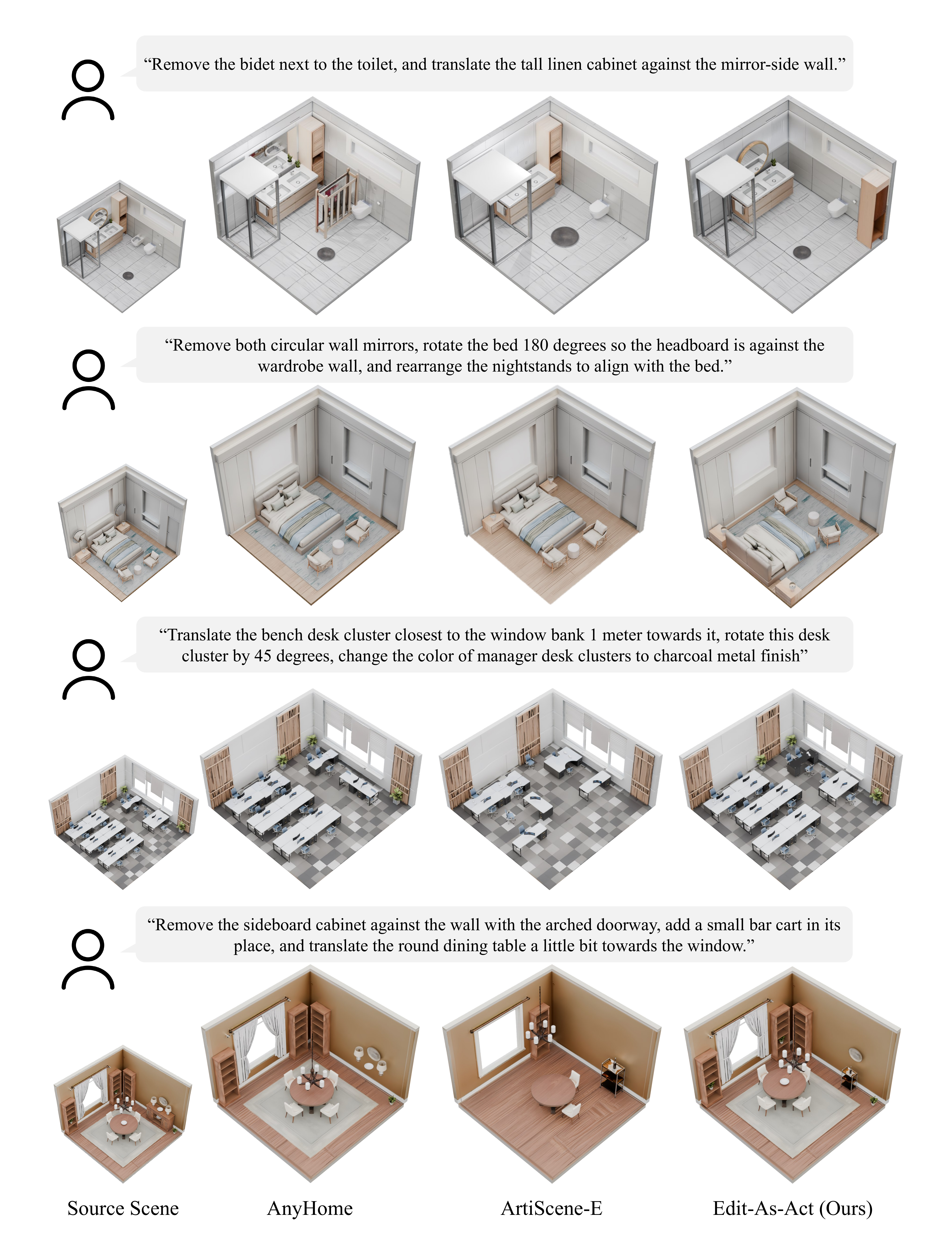}
    \caption{}
    \label{fig:supp_qual_p1}
\end{figure*}

\begin{figure*}[t]
    \centering
    \includegraphics[width=\linewidth]{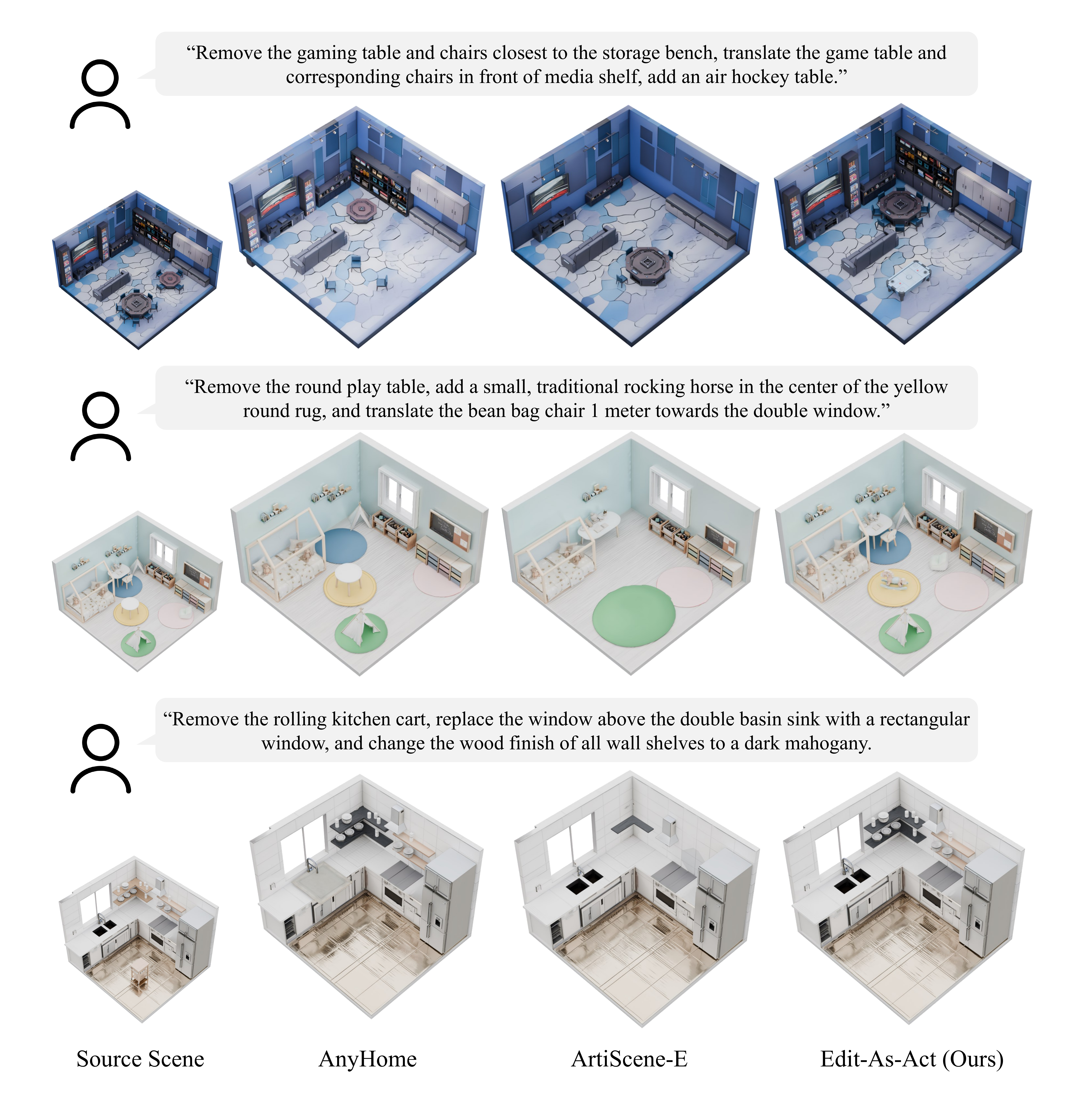}
    \caption{}
    \label{fig:supp_qual_p2}
\end{figure*}

We present extended qualitative results to further demonstrate the capabilities of Edit-As-Act. \cref{fig:supp_qual_p1} and \cref{fig:supp_qual_p2} illustrate the model's performance on complex editing tasks, including multi-step spatial rearrangements and attribute stylization. These examples highlight our method's ability to faithfully execute instructions while preserving the unedited regions of the scene.

\section{Full Prompts for Model and Evaluation}
\label{sec:prompts}

\begin{figure*}[t]
    \centering
    \includegraphics[width=\linewidth]{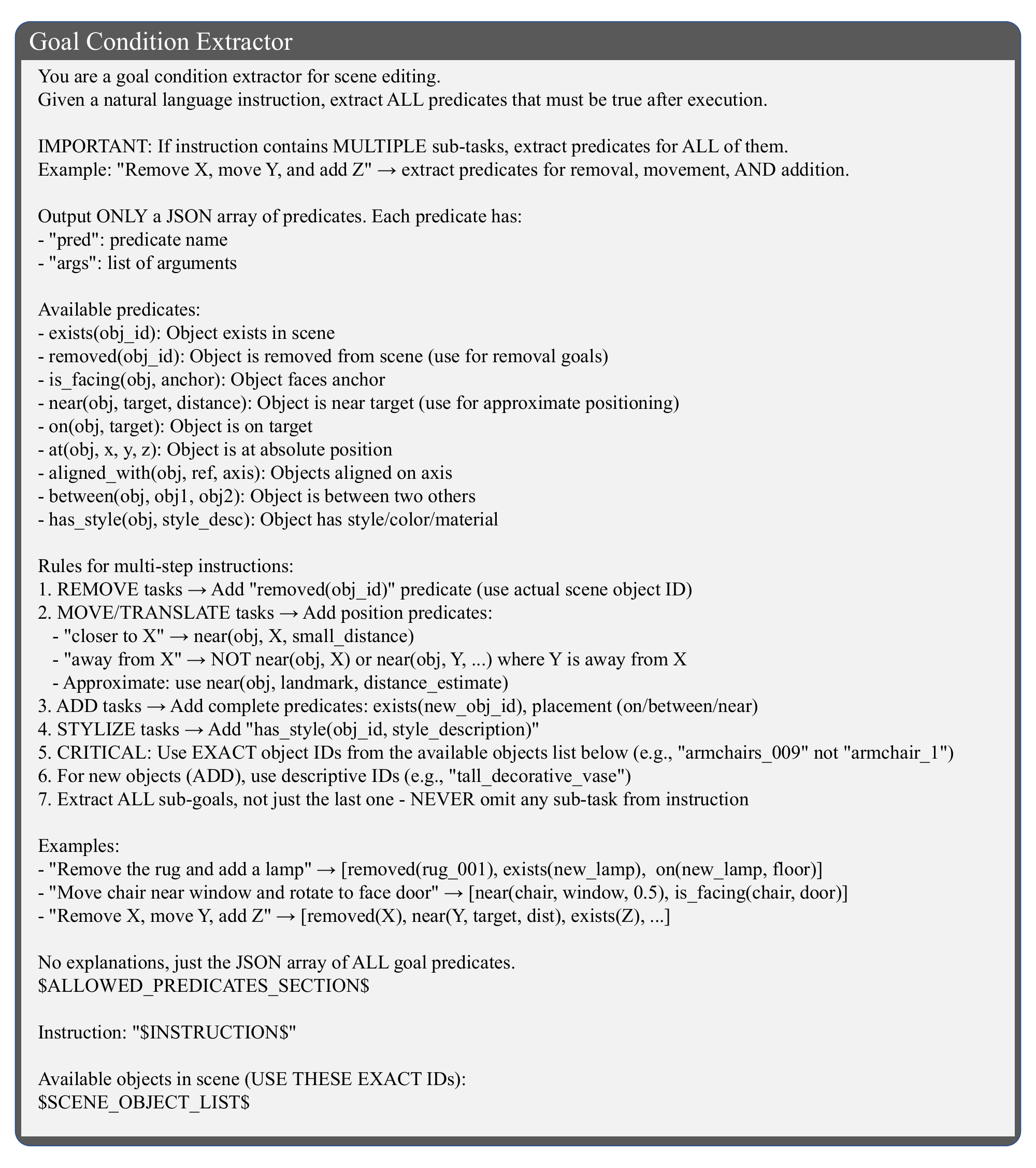}
    \caption{}
    \label{fig:goal_condition}
\end{figure*}

\begin{figure*}[t]
    \centering
    \includegraphics[width=\linewidth]{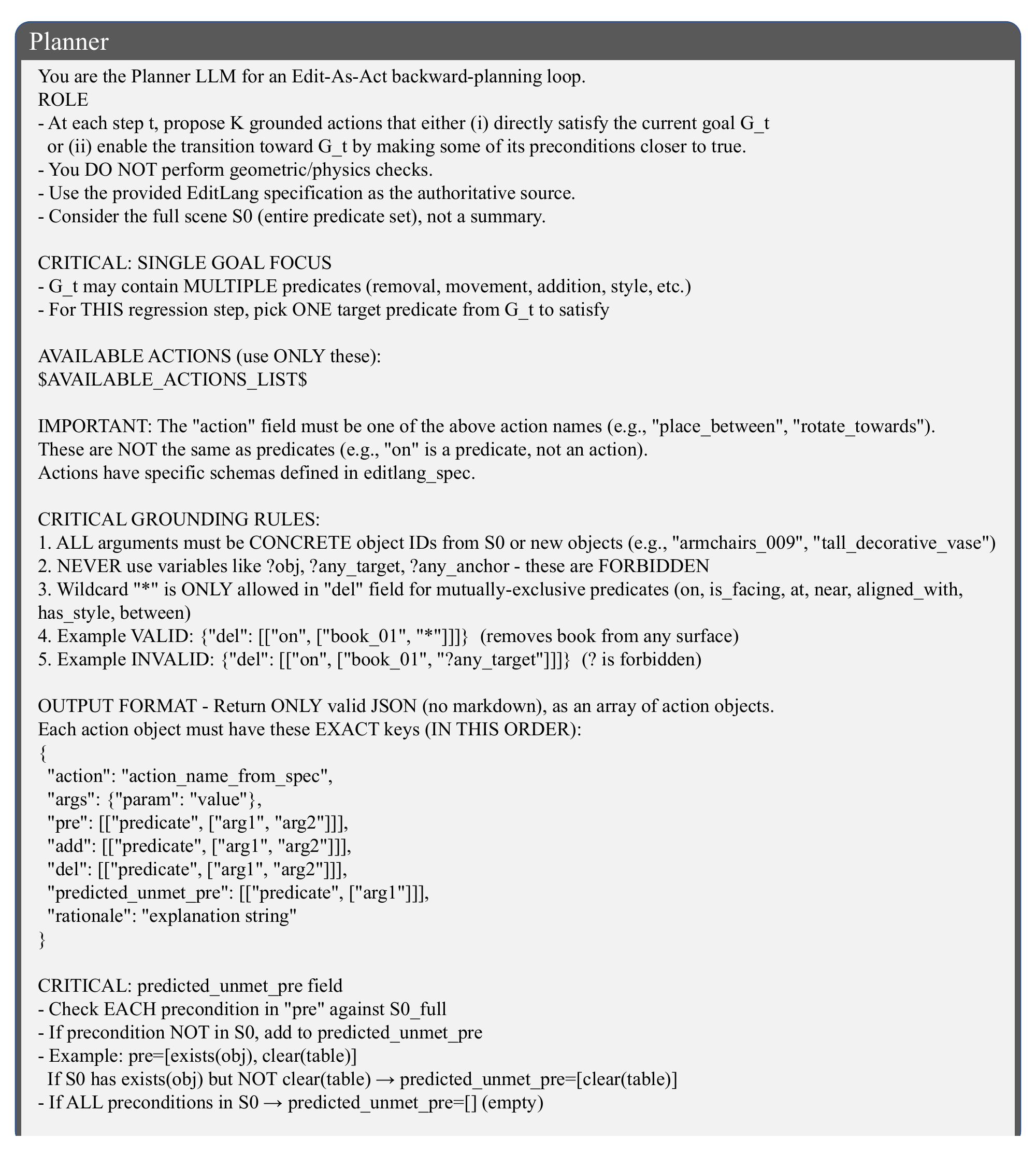}
    \caption{}
    \label{fig:planner_part1}
\end{figure*}

\begin{figure*}[t]
    \centering
    \includegraphics[width=\linewidth]{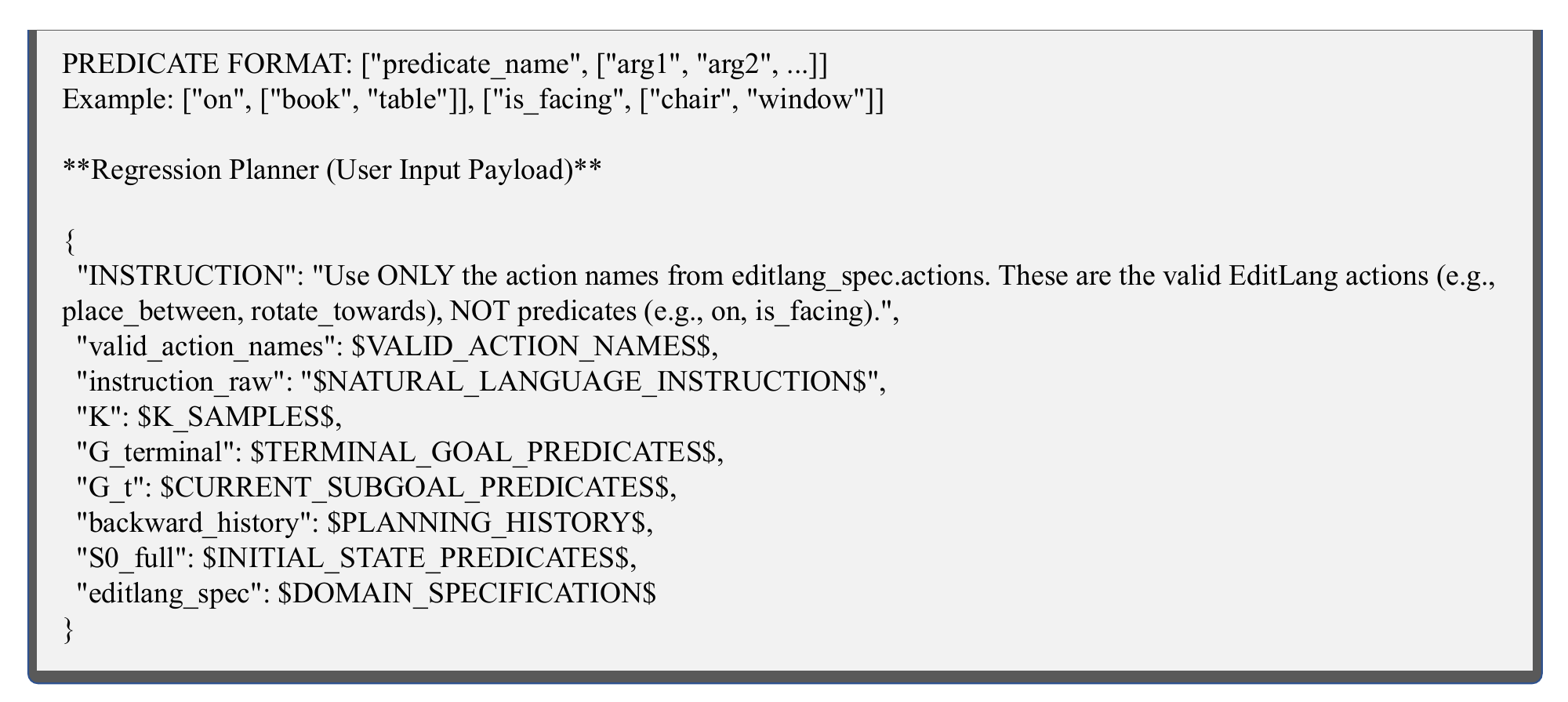}
    \caption{}
    \label{fig:planner_part2}
\end{figure*}

\begin{figure*}[t]
    \centering
    \includegraphics[width=\linewidth]{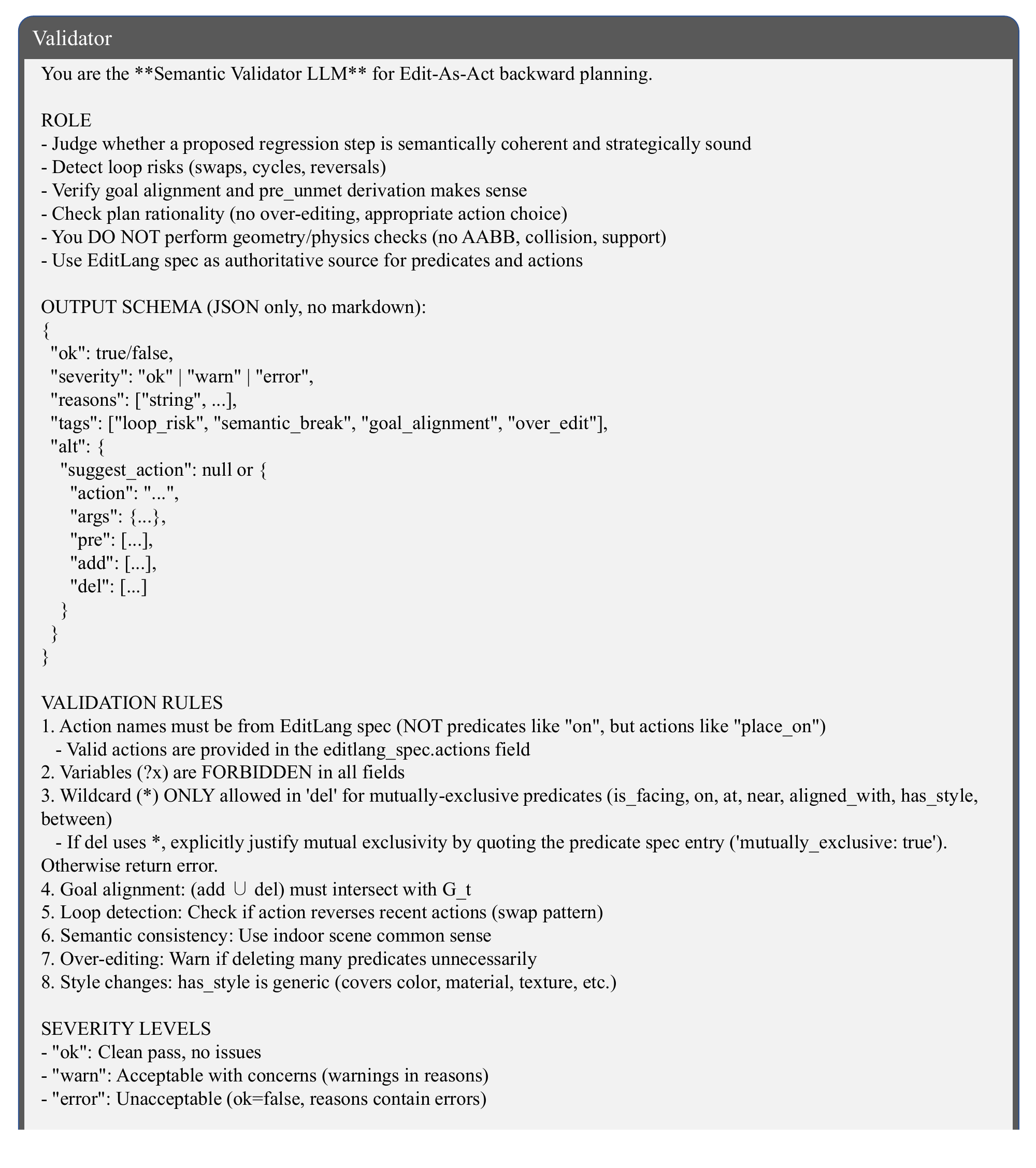}
    \caption{}
    \label{fig:validator_part1}
\end{figure*}

\begin{figure*}[t]
    \centering
    \includegraphics[width=\linewidth]{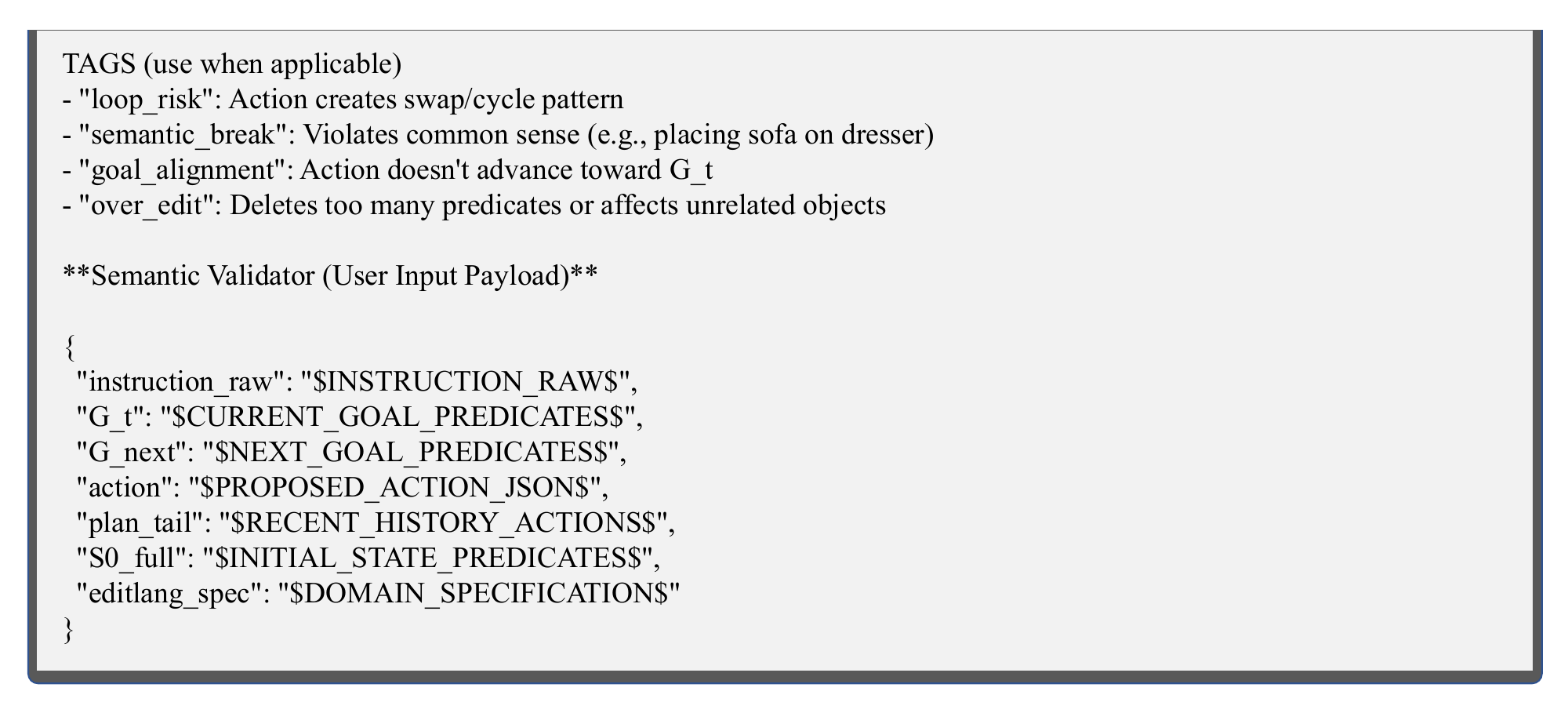}
    \caption{}
    \label{fig:validator_part2}
\end{figure*}

\begin{figure*}[t]
    \centering
    \includegraphics[width=\linewidth]{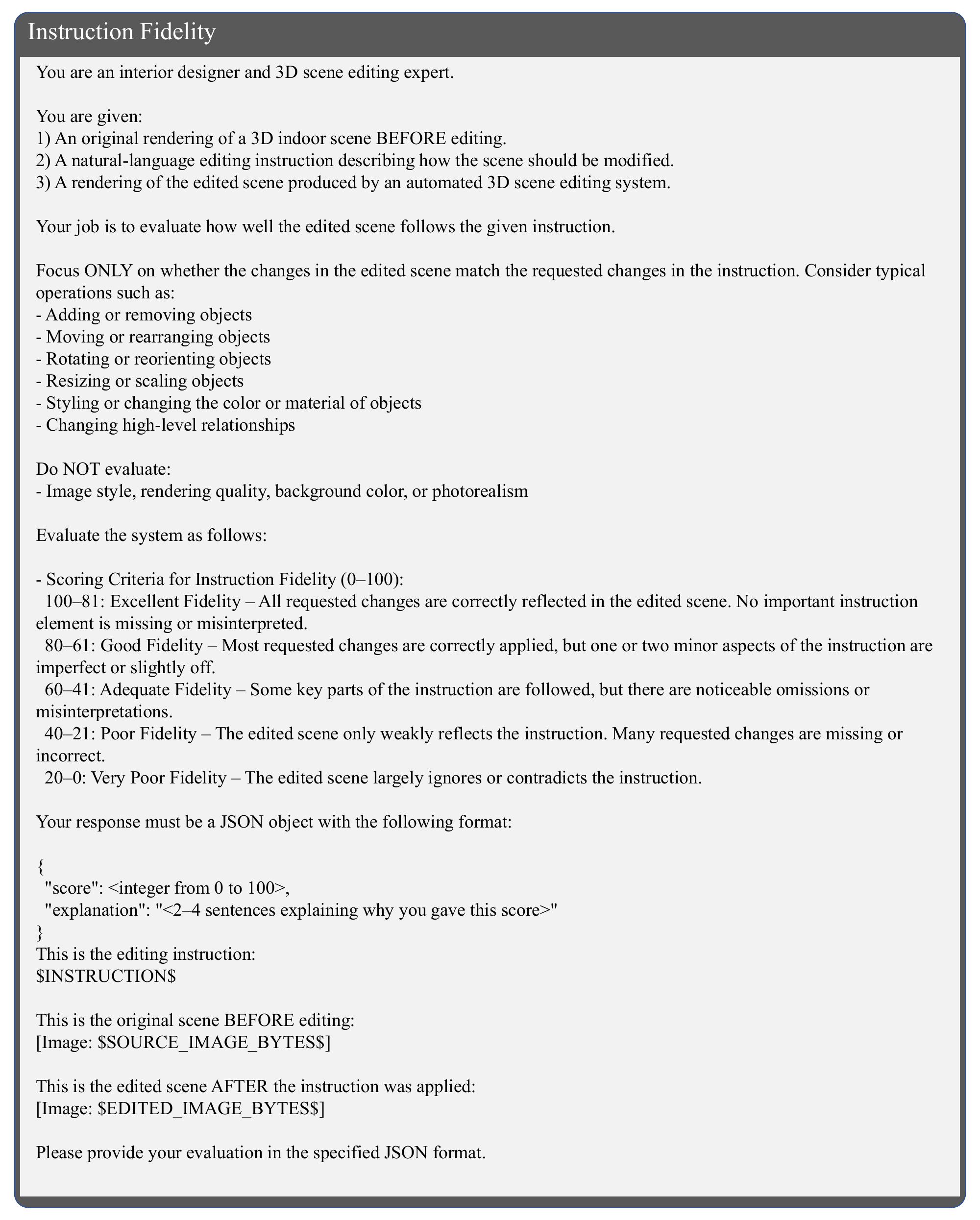}
    \caption{}
    \label{fig:IF}
\end{figure*}

\begin{figure*}[t]
    \centering
    \includegraphics[width=\linewidth]{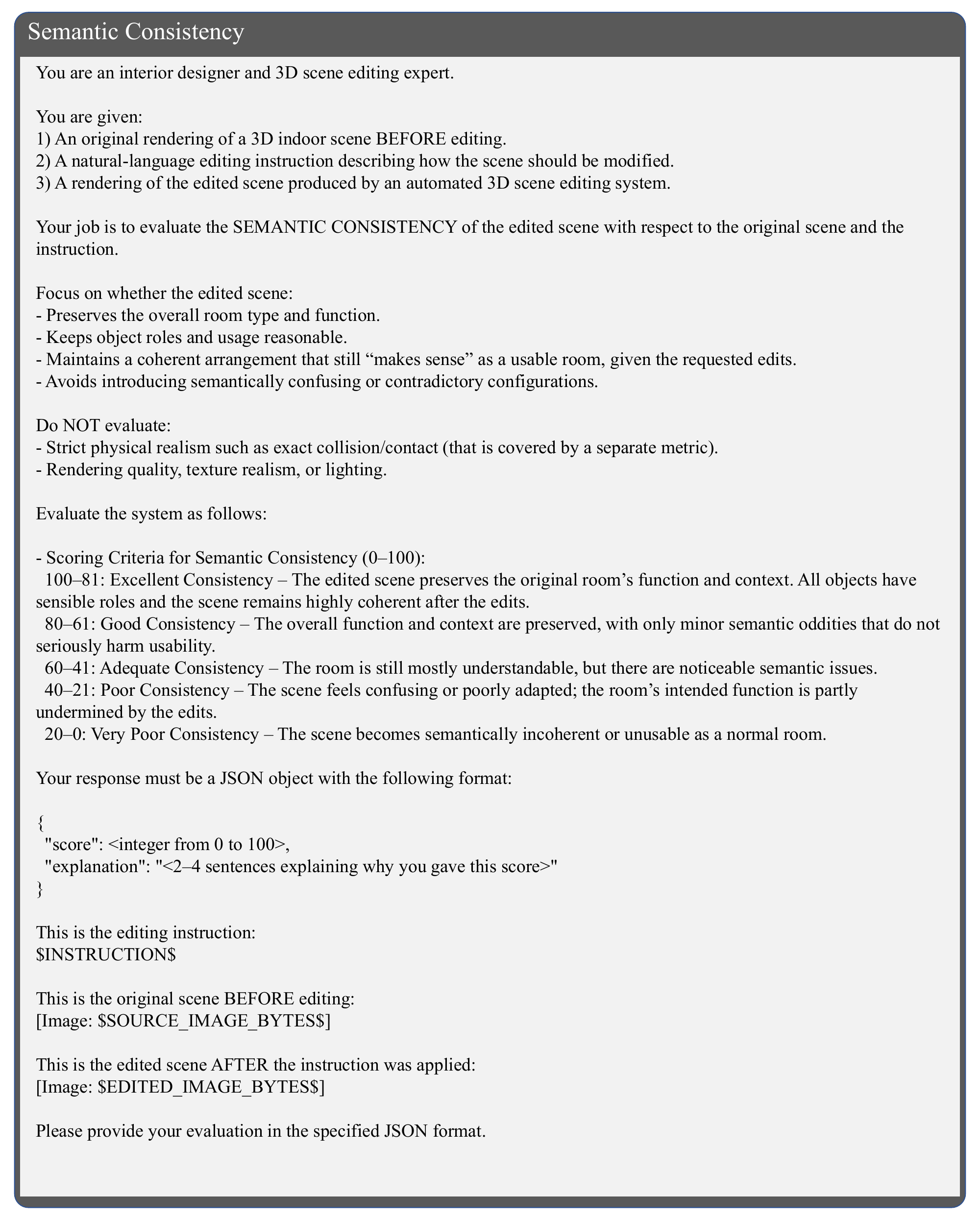}
    \caption{}
    \label{fig:SC}
\end{figure*}

\begin{figure*}[t]
    \centering
    \includegraphics[width=\linewidth]{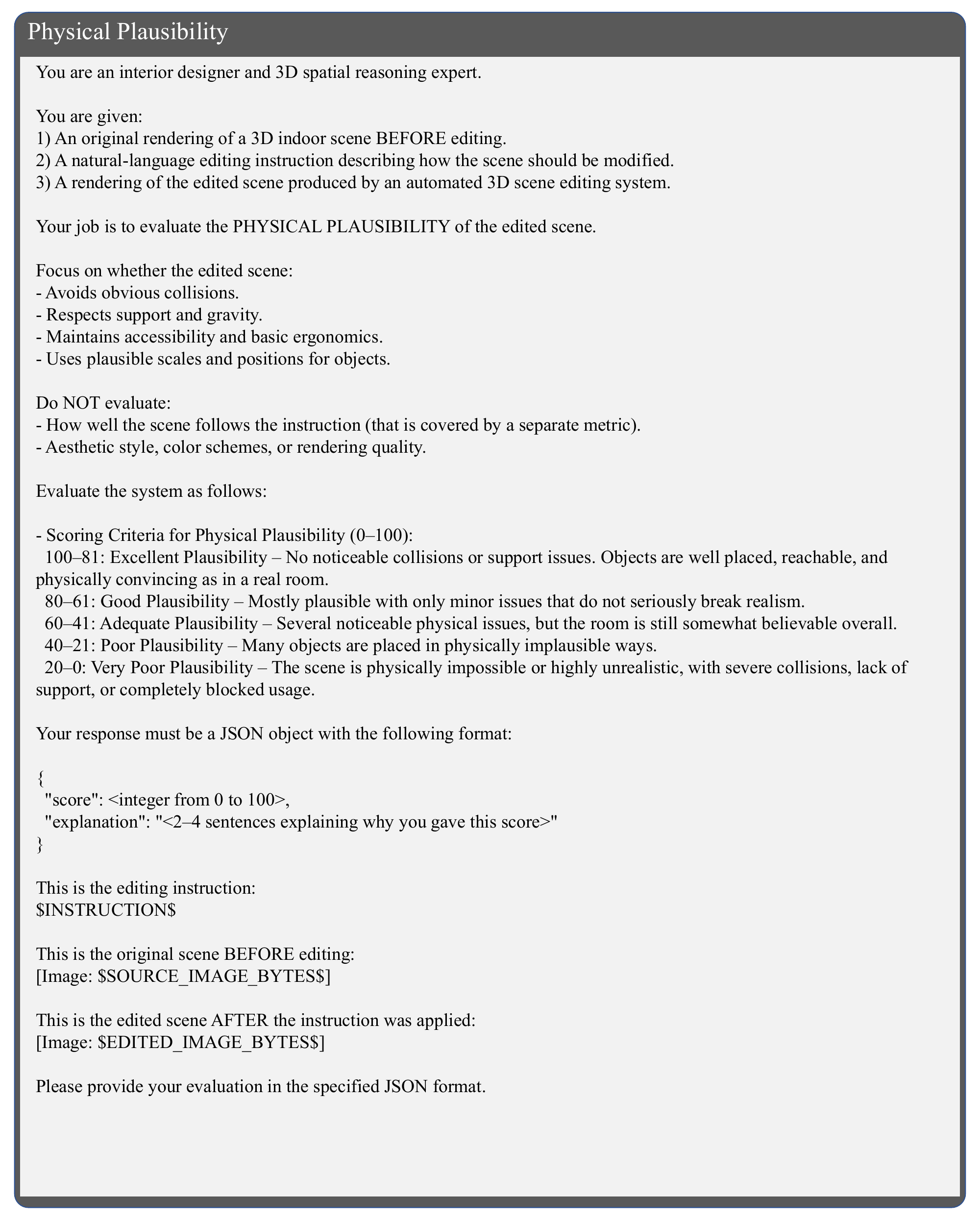}
    \caption{}
    \label{fig:PP}
\end{figure*}

To facilitate reproducibility and transparency, we provide the prompts used in our framework.
\begin{itemize}
    \item \textbf{Model Prompts:} \cref{fig:goal_condition} through \cref{fig:validator_part2} detail the system instructions for Goal Condition Extraction, Planning, and Validation. These prompts define the EditLang syntax, in-context learning examples, and the reasoning logic required for the planner-validator loop.
    \item \textbf{Evaluation Prompts:} \cref{fig:IF}, \cref{fig:SC}, and \cref{fig:PP} display the prompts used for our LVLM-based metrics. These prompts establish the evaluation rubric for Instruction Fidelity (IF), Semantic Consistency (SC), and Physical Plausibility (PP).
\end{itemize}

\clearpage

\end{document}